\documentclass[lettersize,journal]{IEEEtran}
\usepackage{cite}
\usepackage{amsmath,amssymb,amsfonts}
\usepackage{algorithm}
\usepackage{algpseudocode}
\usepackage{graphicx}
\usepackage{textcomp}
\usepackage{wrapfig}
\usepackage{makecell}

\usepackage{array}
\usepackage[caption=false,font=footnotesize]{subfig}
\usepackage{stfloats}
\usepackage{url}
\usepackage{verbatim}
\usepackage{multirow}
\usepackage{tabularx} 
\usepackage{diagbox} 
\usepackage[normalem]{ulem}

\usepackage[colorlinks=true, linkcolor=black, citecolor=black, urlcolor=blue]{hyperref}
\usepackage{wrapfig}
\usepackage{threeparttable}
\usepackage{booktabs}
\usepackage{cleveref}
\crefname{figure}{Fig.}{Figs.}
\DeclareMathOperator*{\argmax}{arg\,max}
\DeclareMathOperator*{\argmin}{arg\,min}
\newcommand{\gravity}{\text{g}^{\text{n}}}
\newcommand{\SO}{\mathrm{SO}(3)}
\newcommand{\dT}{\Delta T}
\NewDocumentCommand{\evalat}{sO{\big}mm}{%
  \IfBooleanTF{#1}
   {\mleft. #3 \mright|_{#4}}
   {#3#2|_{#4}}%
}

\def\BibTeX{{\rm B\kern-.05em{\sc i\kern-.025em b}\kern-.08em
    T\kern-.1667em\lower.7ex\hbox{E}\kern-.125emX}}
\begin{document}
\title{Joint Magnetometer-IMU Calibration via Maximum A Posteriori Estimation}
\author{Chuan Huang, \IEEEmembership{Graduate Student Member, IEEE}, Gustaf Hendeby, \IEEEmembership{Senior Member, IEEE} and Isaac Skog, \IEEEmembership{Senior Member, IEEE}
\thanks{This work has been funded by the Swedish Research Council (Vetenskapsrådet) project 2020-04253 ``Tensor-field based localization". }
\thanks{Chuan Huang is with Dept. of Electrical Engineering and Computer Science,
KTH Royal Institute of Technology (e-mail: chuanh@kth.se).}
\thanks{Gustaf Hendeby is with Dept. of Electrical Engineering, Linköping University (e-mail: gustaf.hendeby@liu.se).}
\thanks{Isaac Skog is with Dept. of Electrical Engineering and Computer Science,
KTH Royal Institute of Technology, and the Div. of Underwater Technology, Swedish Defence Research Agency (FOI), Kista, Sweden (e-mail: skog@kth.se).}}

\maketitle
\begin{abstract}
This paper presents a new method for jointly calibrating a magnetometer and inertial measurement unit (IMU), focusing on balancing calibration accuracy and computational efficiency. The proposed method is based on a maximum a posteriori estimation framework, treating both the calibration parameters and orientation trajectory of the sensors as unknowns. This method enables efficient optimization of the calibration parameters using analytically derived derivatives. The performance of the proposed method is compared against that of two state-of-the-art methods. Simulation results demonstrate that the proposed method achieves the lowest root mean square error in calibration parameters, increasing the calibration accuracy by 20–30\%, while maintaining competitive computational efficiency. Further validation through real-world experiments confirms the practical benefits of the proposed method. The proposed method calibrated 30 magnetometer-IMU pairs in under two minutes on a consumer-grade laptop, which is one order of magnitude faster than the most accurate state-of-the-art algorithm as implemented in this work. Moreover, when calibrated using the proposed method, a magnetic-field-aided inertial navigation system achieved positioning performance comparable to when it is calibrated with the state-of-the-art method. These results demonstrate that the proposed method is a reliable and effective choice for jointly calibrating magnetometer-IMU pairs.
\end{abstract}

\begin{IEEEkeywords}
inertial sensors, magnetometers, in-situ calibration, MAP estimation, IMU preintegration.
\end{IEEEkeywords}
\section{Introduction}
\IEEEPARstart{M}{agnetometers} and inertial measurement units~(IMUs) are widely used in various applications, such as robotics~\cite{pavlasek2023magnetic}, augmented reality~\cite{Liu2023MagLoc}, localization and navigation~\cite{huang2023mains, zmitri2019improving, Joshi2024Enhancing, huang2025inertial, Li2025SCEKFMIO}. However, the accuracy of these sensors, particularly commercial-grade ones, is affected by sensor imperfections such as biases, scale factors, and non-orthogonal sensitivity axes. In addition, magnetometers are highly susceptible to magnetic disturbances, often caused by nearby ferromagnetic materials, further complicating their usability in practical applications.
Therefore, one must calibrate these sensors before use to ensure the best possible performance. While standard IMU and magnetometer calibration methods such as~\cite{Qureshi2017InField,ozyagcilar2012calibrating} address individual sensor errors, physical mounting errors can still introduce coordinate frame misalignment of these sensors. Thus, when multiple sensors are used in a system, cross-sensor calibration becomes essential to ensure proper coordinate system alignment and consistent measurements across all sensors.

When it comes to calibrating low-cost magnetometers and IMUs, in-situ calibration methods, such as those proposed in~\cite{kok2016magnetometer,Zhu2019AnEfficient}, are popular because they do not require additional equipment, and thus save time and cost. Although in-situ calibration methods generally cannot achieve the same accuracy as those that use additional equipment, such as a turntable, they are still sufficient for most use cases involving low- to medium-precision applications. These include pedestrian navigation, smartphone-based localization, and indoor mapping, where sub-degree orientation accuracy and sub-meter positional precision are not strictly required. To that end, in this paper, in situ calibration methods are discussed.

Several in-situ joint magnetometer-IMU calibration methods have been proposed~\cite{kok2016magnetometer, yuanxin2018Dynamic,Papafotis2019MAGICAL, Kok2012MagCal, Dorveaux2009Iterative, Zhu2019AnEfficient}, most of which are based on the assumption that the magnetic field is homogeneous (spatially uniform and temporally stable) and require the sensor platform to be well-exposed to a wide range of orientations. These methods can be broadly categorized into two groups: (1) dynamic system-based methods, which model the temporal evolution of sensor measurements using gyroscope data; (2) static orientation-based methods, which treat sensors' orientations at different timesteps independently. An advantage of dynamic system-based calibration methods is that they allow sensors to be rotated by hand, facilitating easy exposure to a wide range of orientations.
In contrast, static orientation-based methods typically require the sensors to be placed in a more controlled and discrete manner, e.g., by placing them in 15 distinct static orientations \cite{Papafotis2019MAGICAL}. This process can be cumbersome and time-consuming in practice. Moreover, static orientation-based methods often result in limited exposure to the full range of orientations compared to dynamic system-based methods, which continuously incorporate sensor motion and typically expose the sensors to a wider range of orientations.

Building on these advantages, recent research has introduced several dynamic system-based calibration methods. In particular, two representative state-of-the-art dynamic system-based calibration methods are proposed in~\cite{kok2016magnetometer} and \cite{yuanxin2018Dynamic}. In~\cite{kok2016magnetometer}, the magnetometer and IMU calibration problem is formulated as a maximum likelihood (ML) problem, where the likelihood is approximated using an extended Kalman filter~(EKF) and then maximized. This method achieves good calibration accuracy; However, the complexity of the likelihood function makes the derivation of analytic derivatives difficult. As a result, numerical derivatives are used in~\cite{kok2016magnetometer} when maximizing the likelihood, which involves running the EKF multiple times to perform the necessary computations.
In~\cite{yuanxin2018Dynamic}, the calibration problem is formulated as a maximum a posteriori (MAP) estimation problem, where an EKF is used to approximate the posterior probability density of the calibration parameter and current sensor orientation; the full orientation trajectory is not estimated. This method is less computationally demanding; however, its accuracy depends on the quality of the EKF approximation, which is sensitive to the choice of linearization points.

To address the computational complexity of the method in~\cite{kok2016magnetometer} and the sensitivity to linearization points in~\cite{yuanxin2018Dynamic}, we propose a new calibration method that formulates the problem as a joint MAP estimation problem, in which both the full orientation trajectory and calibration parameters are treated as unknowns. This formulation results in a simpler objective function, compared to that in~\cite{kok2016magnetometer}, with closed-form derivatives that facilitate efficient optimization. Furthermore, unlike~\cite{yuanxin2018Dynamic}, our method allows for dynamic adaptation of linearization points during the optimization process. In addition, we analyze the two state-of-the-art methods as well as the proposed method in terms of their underlying assumptions, i.e., the approximations to the probability densities, and give asymptotic upper bounds of the algorithms' computational complexity.
Finally, we evaluate the proposed method and the state-of-the-art methods on both simulation and real-world datasets, highlighting their advantages and key differences.

Our contributions are summarized as follows:
\begin{enumerate}
\item The proposed joint magnetometer-IMU calibration method is fast and accurate, and it can be easily extended to calibrate a magnetometer array and an IMU, which is used in state-of-the-art systems~\cite{zmitri2019improving,huang2023mains,huang2025inertial}. 
\item We establish a connection between our method and two existing state-of-the-art methods by analyzing them within the MAP estimation framework.
\item The empirical evaluation comparing all three methods using both simulated and real-world data helps readers select a suitable method based on their requirements.
\end{enumerate}

\textbf{Reproducible research:}
In the interest of promoting open science and enabling the replication of both our simulation and real-world results, we provide the datasets and algorithms at \href{https://github.com/Huang-Chuan/Mag-IMU-JointCalibration}{https://github.com/Huang-Chuan/Mag-IMU-JointCalibration}.

\section{Problem Formulation}
Consider a sensor system consisting of a three-axis magnetometer and a six-degree-of-freedom IMU. Assume the system is slowly rotated in a homogeneous magnetic field, with minimal translational motion. The dynamics of the corresponding discrete system can be specified as
    \begin{equation}
        x_{k+1} = f(x_k, \tilde{u}_k, \theta,\dT)
    \end{equation}
where the subscript $k$ denotes the time index, $x_k$ denotes the orientation of the IMU frame with respect to the reference frame, $\tilde{u}_k$ denotes the angular velocity measurements, $\theta$ denotes the unknown calibration parameter, and $\dT$ denotes the sampling interval. The calibration parameter accounts for all variables contributing to systematic measurement errors, such as sensor bias, as well as environmental factors, such as magnetic inclination. The exact quantities used in calibration are specified based on the users' needs.

The measurements from an accelerometer and a magnetometer, as well as the measurements from a gyroscope, can be modeled as
\begin{subequations}  \label{eq: sensor measurement model}
    \begin{align}
    y_k & \triangleq \begin{bmatrix}
        \tilde{s}_k \\
        \tilde{m}_k
    \end{bmatrix} = \begin{bmatrix}
        h^a(x_k;\theta) + e_k^{a}\\
        h^m(x_k;\theta) + e_k^{m}\\
    \end{bmatrix}  \label{eq: accelerometer and gyroscope measurement model}\\
     \intertext{and}
    \tilde{u}_k &= h^\omega(u_k;\theta) + e_k^{\omega}
\end{align}
\end{subequations}
respectively.
Here, $\tilde{s}_k\in \mathbb{R}^3$ and $\tilde{m}_k \in \mathbb{R}^3$ denote the accelerometer and magnetometer measurement, respectively. Moreover, $u_k \in \mathbb{R}^3$ denotes the true angular velocity, $e_k^{(\cdot)} \in \mathbb{R}^3$ denotes the additive white noise associated with each type of sensor, and its covariance is specified by $\Sigma_{(\cdot)}$. Further, the noise terms are assumed to be independent of each other.

Let \text{$y_{1:T} \!\triangleq\! (y_1, y_2, \ldots, y_T)$} denote the sequence of accelerometer and magnetometer measurements from time 1 to \( T \), and \text{$\tilde{u}_{0:{T-1}} \!\triangleq\! (\tilde{u}_0,\tilde{u}_1,\ldots,\tilde{u}_{T-1})$} denote the sequence of angular velocity measurements from time 0 to \( T-1 \). The purpose of calibration is to estimate $\theta$ given the measurements \text{$y_{1:T}$} and \text{$\tilde{u}_{0:{T-1}}$}. 
\section{Proposed Method and its Connection to State-of-the-Art}
In this section, two related yet distinct approaches to estimate $\theta$ are discussed. The first employs a MAP estimator based on the joint posterior distribution of the orientation trajectory $x_{0:T}$ and calibration parameters, and forms the core of the proposed calibration method. The second employs a MAP estimator based on the marginal posterior distribution of the calibration parameters, giving rise to the calibration methods presented in \cite{kok2016magnetometer} and \cite{yuanxin2018Dynamic}. The aim of this section is to show the connection between the proposed method and existing state-of-the-art methods.

\subsection{MAP Estimation Using the Joint Posterior} \label{subsec:joint posterior}
A MAP estimator can be used to estimate the calibration parameter $\theta$ and the orientation trajectory $x_{0:T}$ jointly. Formally, the estimate is given by~\cite[p.15]{Simon2013Bayesian}  
\begin{subequations} \label{eq: joint MAP estimator}
    \begin{align} 
    (\hat{\theta}, \hat{x}_{0:T})^{\text{\tiny{MAP}}} &\triangleq \argmax_{\theta,x_{0:T}} p(\theta, x_{0:T} \!\mid\! y_{1:T}, \tilde{u}_{0:T-1}) \label{eq: full map estimator} \\
    &= \argmax_{\theta,x_{0:T}} \Bigl( p(y_{1:T}\! \mid \!\theta, x_{0:T}, \tilde{u}_{0:T-1}) \nonumber\\ 
    &\quad\quad\times p(x_{0:T}, \theta \!\mid\! \tilde{u}_{0:T-1}) \Bigr). \label{eq: Bayesian full map estimator}
    \end{align}
\end{subequations}
Here, \text{$p(A\!\mid\! B)$} denotes the conditional probability density function of the random variable $A$ given the random variable $B$. Equation~\eqref{eq: Bayesian full map estimator} follows directly from Bayes’ rule applied to~\eqref{eq: full map estimator}, omitting the normalizing factor \text{$p(\tilde{u}_{0:T-1} \!\mid \!y_{1:T})$}. Furthermore, the terms \text{$p(y_{1:T} \!\mid \!\theta, x_{0:T}, \tilde{u}_{0:T-1})$} and \text{$p(x_{0:T}, \theta \!\mid \!\tilde{u}_{0:T-1})$} in~\eqref{eq: Bayesian full map estimator} can be factorized as
\begin{subequations}
    \begin{align}
        &p(y_{1:T} \mid \theta, x_{0:T}, \tilde{u}_{0:T-1}) = \prod_{k=1}^T p(y_k \mid \theta, x_{k})\label{eq: likelihood}\\
        \intertext{and}
        &p(x_{0:T}, \theta\mid \tilde{u}_{0:T-1}) = p(\theta) p(x_0)  \prod_{k=0}^{T-1} p(x_{k+1}|x_k,\theta,\tilde{u}_k) \label{eq: dynamics}
    \end{align}    
\end{subequations}
respectively. Here, $p(\theta)$ and $p(x_0)$ denote the priors on the parameters and initial orientation, respectively. Furthermore, $p(y_k\mid \theta,x_k)$ and  $p(x_{k+1}\mid x_k, \theta,\tilde{u} _k)$ denote the measurement likelihood and Markov state transition probability of the orientation dynamics, respectively.

Equations~\eqref{eq: joint MAP estimator},~\eqref{eq: likelihood}, and~\eqref{eq: dynamics} form the foundation of the proposed method. How to practically solve the optimization problem in~\eqref{eq: joint MAP estimator} will be discussed in Section~\ref{sec: implementation}.

\subsection{MAP Estimation Using the Marginal Posterior}
A MAP estimator can also be used to estimate the calibration parameter \textit{only}. This is done by maximizing the marginal posterior probability density of the calibration parameter $\theta$, which is given by~\cite[p.18]{Simon2013Bayesian}
\begin{subequations}
\begin{equation}  \label{eq: marginalized MAP}
    \hat{\theta}^{\text{\tiny{MAP}}} =  \argmax_{\theta} p(\theta \mid y_{1:T}, \tilde{u}_{0:T-1})
\end{equation}
where 
\begin{equation}  \label{eq: marginization}
     p(\theta \mid y_{1:T}, \tilde{u}_{0:T-1}) = \int p(\theta, x_{0:T} \mid y_{1:T}, \tilde{u}_{0:T-1})\;d x_{0:T}.
\end{equation}
\end{subequations}
Note the integrand in~\eqref{eq: marginization} is exactly the probability density that is to be maximized in \eqref{eq: full map estimator}.  

The maximization in~\eqref{eq: marginalized MAP} can be equivalently written as \cite{kok2016magnetometer}
\begin{subequations}
\begin{equation} 
      \hat{\theta}^{\text{\tiny{MAP}}} = \argmax_{\theta} p(\theta)\prod_{k=1}^{T} p(y_k\mid y_{1:k-1}, \theta, \tilde{u}_{0:k-1})\label{eq: manon ML equation}
\end{equation}
where 
\begin{equation}
    p(y_1\!\mid\! y_{1:0}, \theta, \tilde{u}_0)\!\triangleq\! p(y_1 \!\mid\! \theta, \tilde{u}_0).
\end{equation}
\end{subequations}
In~\cite{kok2016magnetometer} by Kok et al., the marginal likelihood $p(y_k\!\mid \!y_{1:k-1}, \theta, \tilde{u}_{0:k-1})$ is approximated by the one-step ahead measurement prediction density in an EKF. In addition, the prior $p(\theta)$ is assumed to be uniformly distributed; thus, the estimator used by Kok et al. is equivalent to an ML estimator.

Alternatively,~\eqref{eq: marginalized MAP} can also be equivalently expressed as
\begin{equation} \label{eq: Wu's}
    \hat{\theta}^{\text{\tiny{MAP}}} =  \argmax_{\theta} \int p(\theta, x_T \mid y_{1:T}, \tilde{u}_{0:T-1}) dx_T 
\end{equation}
which reflects what is done in the calibration method proposed in~\cite{yuanxin2018Dynamic} by Wu et al.. They used an EKF with an augmented state-space model
\begin{subequations} \label{eq: augmented state-space model}
    \begin{align}
    x_{k+1} &= 
     f(x_k, \tilde{u}_k, \theta_k, \Delta T) \\
   \theta_{k+1} &= \theta_k
\end{align}
\end{subequations}
to compute an approximation to the integrand in~\eqref{eq: Wu's}. Since $p(\theta, x_T \mid y_{1:T}, \tilde{u}_{0:T-1})$ is approximated as a joint Gaussian distribution density, the marginal posterior $p(\theta \mid y_{1:T}, \tilde{u}_{0:T-1})$ is also a Gaussian distribution density. Further, given the Gaussian distribution, the maximization process in~\eqref{eq: marginalized MAP} corresponds to selecting the mean of the marginal posterior.

\subsection{Comparison of Calibration Methods} \label{discussion}

The first approach using the joint posterior in Section~\ref{subsec:joint posterior} offers the proposed calibration method several advantages, notably its minimized dependence on probability distribution approximations and its favorable computational efficiency.

\subsubsection{Accuracy} 
Within the first approach, all probability terms can be evaluated exactly under Gaussian noise assumptions, with the exception of the transition term \text{$p(x_{k+1}\!\mid\! x_k,\theta, \tilde{u}_k)$}. However, as discussed in~\cite{forster2016manifold}, this transition term can be accurately approximated given reasonably small gyroscope noise.

More importantly, the estimator does not assume a specific joint probability distribution of the calibration parameter and orientations, nor does it depend on the Gaussian approximations inherent to an EKF. By avoiding such assumptions and approximations, the proposed method preserves the true posterior structure and can potentially yield more accurate estimates.

In contrast, the methods in~\cite{kok2016magnetometer} and~\cite{yuanxin2018Dynamic} adopt the second approach. They utilized an EKF to approximate the marginal likelihood or posterior density. Consequently, their accuracy is vulnerable to factors such as linearization point selection, filter consistency, and the validity of Gaussian assumptions.

\subsubsection{Computational Efficiency} When the priors \text{$p(\theta)$} and \text{$p(x_0)$} are Gaussian distributed or uninformative, i.e., \(p(\cdot) \propto 1\), the MAP estimation problem in~\eqref{eq: joint MAP estimator} reduces to a nonlinear least squares problem, which can be efficiently solved using standard solvers such as Gauss–Newton or Levenberg–Marquardt. Due to the sparsity of the Jacobian matrices associated with the optimization problem, which is similar to those encountered in the exploration tasks during the simultaneous localization and mapping process~\cite{Kaess2008iSAM}, the computational complexity per iteration is $\mathcal{O}(3T+\text{dim}(\theta))$. The factor 3 corresponds to the three degrees of freedom in the orientation representation for each of the $T$ poses. Assuming convergence in $N_{\text{iter}}$ iterations, the total computational complexity of the proposed method is $\mathcal{O}\left(N_{\text{iter}} (3T+\text{dim}(\theta)) \right)$.

In the method~\cite{kok2016magnetometer}, a quasi-Newton optimization solver is used to maximize the marginal likelihood function, where numerical derivatives are used due to the difficulty of deriving closed-form Jacobians. The numerical derivatives needed by the optimization program require running an EKF $\text{dim}(\theta)$ times. Given an EKF state dimension of $d=3$, the computational complexity per time step is $\mathcal{O}(d^3)$~\cite{Daum2005Nonlinear}. Consequently, the total complexity for~\cite{kok2016magnetometer} is $\mathcal{O}(N_{\text{iter}}\times \text{dim}(\theta)\times T \times  3^3)$, where $N_{\text{iter}}$ is the number of iterations to convergence. In practice, the number of iterations $N_{\text{iter}}$ required for convergence in this method tends to be significantly larger than that of the proposed calibration method. Further, the complexity in~\cite{kok2016magnetometer} scales as a product of $\dim(\theta)$ and $T$, whereas our method scales as a sum. These two key differences may explain the slow performance of the method in~\cite{kok2016magnetometer} in the experiments. 

Alternatively, the method in~\cite{yuanxin2018Dynamic} avoids numerical derivatives by running an EKF filter with the augmented state-space model~\eqref{eq: augmented state-space model}. This leads to a computational complexity of $\mathcal{O}\left(T \times(\text{dim}(\theta) + 3)^3 \right)$. Compared to its optimization-based counterparts, this method is computationally more efficient, especially when $\text{dim}(\theta)$ is small compared to $N_{\text{iter}}$.

\section{Implementation} \label{sec: implementation}
Next, we present the state-space model employed for joint magnetometer-IMU calibration, along with the implementation details of the proposed method.
\subsection{State-Space Model and Probabilities}
\begin{figure}[tb!]
    \centering
    \includegraphics{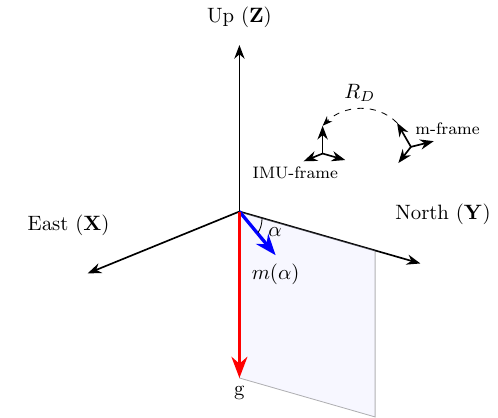}
    \caption{The reference (big axes sets) and sensors' coordinate frames (small axes sets). The Z-axis in the reference frame is aligned with the local gravity vector, $\text{g}$, the Y--Z plane is parallel to the local magnetic field, $m(\alpha)$, with its horizontal component pointing in the positive Y-axis direction. The dip angle of the magnetic field $\alpha$ is the angle between the magnetic field and the horizontal plane. The misalignment of the magnetometer frame (m-frame) with the inertial sensor frame (IMU-frame) is represented by the rotation matrix $R_D$.}
    \label{fig: coordinate frame}
\end{figure}

 The unknown calibration parameter $\theta$ is defined as
\begin{equation} \label{eq: theta specification}
    \theta \triangleq (o^a, o^\omega, D^m, o^m, \alpha)
\end{equation}
where $o^a \in \mathbb{R}^3$, $o^\omega \in \mathbb{R}^3$, and $o^m \in \mathbb{R}^3$ denote the accelerometer, gyroscope, and magnetometer biases, respectively. Moreover, $D^m \in \mathbb{R}^{3\times 3}$ denotes the magnetometer distortion matrix, and $\alpha \in \mathbb{R}$ denotes the dip angle of the local magnetic field, as shown in~Fig. \ref{fig: coordinate frame}. The orientation $x_k$ is parameterized as a rotation matrix $R_k \in \SO$. In the sequel, $x_k$ is used interchangeably with $R_k$. 

Assume that the true angular velocity remains approximately constant over the sampling interval and the gyroscope measurements are affected by additive biases, i.e., \text{$h^\omega(u_k;\theta)= u_k + o^\omega$}. The system dynamics are then described by
\begin{subequations}
\begin{align}
\label{eq: dynamic model with noisy input}
    f(x_k, \tilde{u}_k, \theta,\dT) &=R_k \text{Exp} \big(  (\tilde{u}_k - o^\omega - e_k^\omega) \dT \big)\\
    \intertext{where}
    \text{Exp}(v) &\triangleq e^{v^{\wedge}}, \quad v\in \mathbb{R}^3\\
    v^{\wedge} &\triangleq \begin{bmatrix}
        0 & -v_z& v_y\\
        v_z & 0 & -v_x\\
        -v_y & v_x & 0
    \end{bmatrix}.
\end{align}
\end{subequations}

The sensors' measurement functions in~\eqref{eq: accelerometer and gyroscope measurement model} are~\cite{kok2016magnetometer}
\begin{subequations} \label{eq: actual sensor measurement model}
    \begin{align}
    \label{eq: accelerometer model}
    h^a(x_k;\theta) &= -R_k^{\top} \text{g}^{\text{n}} + o^a \\
    \label{eq: magnetometer model}
    h^m(x_k;\theta) &= D^m  R_k^{\top} m^{\text{n}}(\alpha) + o^m.
\end{align}
\end{subequations}
Here $\text{g}^{\text{n}} = \left[0 \; 0 \; \text{g}_0\right]^{\top}$ and $m^{\text{n}}(\alpha) = \big[0 \; \text{cos}(\alpha) \; -\text{sin}(\alpha)\big]^{\top}$ denote the local gravity and the normalized local magnetic field expressed in the reference frame, respectively. $\text{g}_0$ is the gravitational acceleration constant. Note that the distortion matrix $D^m$ can 
be factorized as 
\begin{equation} \label{eq: factorized distortion matrix}
    D^m = D^{I} R_{D}
\end{equation}
where $D^I \in \mathbb{R}^{3\times3}$ denotes the intrinsic calibration matrix and $R_D \in \SO$ denotes the rotation matrix that aligns the magnetometer frame with the inertial sensor frame. 

The equations in~\eqref{eq: actual sensor measurement model} allow us to specify the likelihood in~\eqref{eq: likelihood} as
\begin{align} \label{eq: measurement probability densities}
    p(y_k\mid x_k,\theta) &= \mathcal{N}\left( y_k \Bigg| \begin{bmatrix}
        h^a(x_k;\theta)\\
        h^m(x_k;\theta)
    \end{bmatrix}, \begin{bmatrix}
        \Sigma_a & 0\\
        0 & \Sigma_m
    \end{bmatrix} \right).
\end{align}
Here $\mathcal{N}(\mu, P)$ denotes a Gaussian distribution with mean $\mu$ and covariance $P$, and $\mathcal{N}(\cdot \mid\mu, P)$ is the corresponding probability density function. Further,~\eqref{eq: dynamic model with noisy input} specify the Markov state transition probability densities \text{$p(x_{k+1}\!\mid\! x_k, \tilde{u}_k, \theta)$} in~\eqref{eq: dynamics} as
\begin{subequations}
    \label{eq: uncertainty in SO3}
    \begin{align}
    \label{eq: uncertainty in SO3 (a)}
        p(x_{k+1}\mid x_k, \tilde{u}_k, \theta) &\approx \frac{1}{\sqrt{\text{det}(2\pi\Sigma_w)}} e^{-\frac{1}{2} \|\delta u_k\|_{\Sigma_w}^2} \\ 
        \delta u_k & \triangleq \tilde{u}_k - \frac{1}{\dT} \text{Log}(R_k^{\top} R_{k+1})- o^\omega \\
        \text{Log}(R) &\triangleq a \psi,\; a \in \mathbb{R}^3, \psi \in \mathbb{R}.
    \end{align}    
\end{subequations}
Here $a$ and $\psi$ denote the rotation axis and the rotation angle of $R$, respectively. The definition of the operation $\text{Log}(\cdot)$ can be found in~\cite{forster2016manifold}. Further, $\text{det}(\cdot)$ denotes the determinant of a matrix, and \text{$\| \cdot \|_{\Sigma}^2 \triangleq (\cdot)^{\top} \Sigma^{-1} (\cdot)$}. The approximative density function~\eqref{eq: uncertainty in SO3 (a)} is the result of applying the theorem in~\cite{forster2016manifold}, which states if $R\in \SO$ and $\tilde{R} \in \SO$, and they are related via $\tilde{R} = R \,\text{Exp}(v)$, where $v\sim \mathcal{N}(0, \Sigma)$ and $\Sigma$ is sufficiently small,
then $p(\tilde{R}) \approx \frac{1}{\sqrt{ \det(2\pi\Sigma)}} e^{-\frac{1}{2} \|\text{Log}(R^{\top}\tilde{R})\|_{\Sigma}^2}$.

\subsection{Cost Functions}
Since the logarithm $\ln(\cdot)$ is a monotonic function, the MAP estimator in \eqref{eq: joint MAP estimator} can be equivalently obtained by minimizing the cost function $L(\theta,x_{0:T})$, i.e.,
\begin{subequations} \label{eq: original MAP}
    \begin{equation} 
        (\hat{\theta}, \hat{x}_{0:T})^{\text{\tiny{MAP}}} \triangleq \argmin_{\theta, x_{0:T}} L(\theta,x_{0:T})      
    \end{equation}
where
    \begin{equation}
        \begin{split} 
            L(\theta,x_{0:T})=&-\ln(p(\theta))-\ln(p(x_0)) \\ &+\sum_k\|\tilde{s}_k+R_k^{\top} \text{g}^{\text{n}} - o^a\|^2_{\Sigma_a}\\
            & +  \sum_k\|\tilde{m}_k- D^m R_k^{\top} m(\alpha) - o^m\|^2_{\Sigma_m} \\
            &+  \sum_k\|  \tilde{u}_k - \frac{1}{\dT} \text{Log}(R_k^{\top} R_{k+1} ) - o^\omega \|^2_{\Sigma_\omega}.
        \end{split}
    \end{equation}
\end{subequations}
Note that the cost function $L(\theta,x_{0:T})$ is simply the negative logarithm of the joint posterior density in~\eqref{eq: joint MAP estimator}. The first two terms are prior terms, and the last three terms correspond to~\eqref{eq: measurement probability densities} and~\eqref{eq: uncertainty in SO3}. 

In cases where the sample rate of the magnetometer is $N$ times slower than that of the IMU sensor due to hardware limitations or downsampling to reduce computational load, the cost function can be modified accordingly, i.e.,
\begin{subequations}\label{eq: MAP preintegration}
\begin{equation}
            (\hat{\theta}, \hat{x}_{0:T})^{\text{\tiny{MAP}}} \triangleq \argmin_{\theta, x_{0:T}} L(\theta,x_{0:T}^{(N)})     
\end{equation}

\begin{equation}\label{eq: preintegration cost}
\begin{split} 
    L(\theta,x_{0:T}^{(N)})=&-\ln(p(\theta))-\ln(p(x_0)) \\
    & +  \sum_{k^\prime}\|\tilde{s}_{k^\prime}+R_{k^\prime}^{\top} \text{g}^{\text{n}} - o^a\|^2_{\Sigma_a}\\
    & +  \sum_{k^\prime}\|\tilde{m}_{k^\prime}- D^m R_{k^\prime}^{\top} m(\alpha) - o^m\|^2_{\Sigma_m} \\
    & +  \sum_{k^\prime}\| \text{Log}(\Delta \tilde{R}_{k^\prime, k^\prime+N}^{\top} R_{k^\prime}^{\top} R_{k^\prime+N} )\|^2_{\Sigma_{\Delta \tilde{R}_{k',k'+N}}}
\end{split}
\end{equation}
where 
\begin{align}
    x_{0:T}^{(N)} &= \{x_0, x_{N-1}, x_{2(N-1)}, \ldots \}\\ 
    k' &= (N - 1)(l -1), \; l=1,2,\ldots\\
    \Delta \tilde{R}_{i,j} &= \prod_{k=i}^{j-1} \text{Exp}\big((\tilde{u}_k - o^\omega) \dT \big) \label{eq: preintegration}\\ 
    \Sigma_{\Delta\tilde{R}_{k',k'+N}} &= \sum_{k=k'}^{k'+N-1} A_k \Sigma_w A_k^{\top}\\
    A_k  &\triangleq  \Delta \tilde{R}_{k+1, k'+N - 1}^{\top}J_k^r \dT \\
    J_k^r &\triangleq J^r \big((\tilde{u}_k - o^\omega) \dT \big).
\end{align}
\end{subequations}
Here $J^r(\cdot) \in \mathbb{R}^3$ is the right Jacobian of $\SO$ and it is given by~\cite[p.~40]{chirikjian2011stochastic}
\begin{equation}
    J^r(v)= I - \frac{1-\text{cos}(\|v\|)}{\|v\|^2} v^\wedge + \frac{\|v\|-\text{sin}(\|v\|)}{\|v\|^3} (v^\wedge)^2.
\end{equation}
Note the terms in the last sum in~\eqref{eq: preintegration cost} are related to the probability density 
\begin{equation}
    p(x_{k'+N}\!\mid \!x_{k'}, \theta, \tilde{u}_{k'}, \ldots, \tilde{u}_{k'+N-1}) \! = \!\!\!\prod_{k=k'}^{k'+N -1 } \!\!\!p(x_{k+1}|x_k,\theta,\tilde{u}_k)
\end{equation}
which is given by IMU preintegration proposed in~\cite[Eq.~38]{forster2016manifold}.

\subsection{Optimization on Manifolds}
When the priors on $\theta$ and $x_0$ are Gaussian or uninformative, the optimization problems in~\eqref{eq: original MAP} or~\eqref{eq: MAP preintegration} are reduced to nonlinear least squares problems, which can be solved by the Gauss-Newton algorithm~\cite{Anders2023Optimization}.
However, since orientation matrices live on the manifold $\SO$, special treatment is required. In particular, the optimization problem is solved in the tangent space of $\SO$, and interested readers are referred to~\cite{forster2016manifold} for more details. A brief review of the Gauss-Newton algorithm on manifolds and the analytic derivatives required by the solver are given in the Appendix~\ref{app:GN},~\ref{app:first}, and~\ref{app:second}. Readers may also opt for more robust algorithms, such as the Levenberg–Marquardt method \cite[Ch. 6]{Anders2023Optimization}, to handle ill-conditioned Hessian matrices when necessary. Moreover, if good prior knowledge of $\theta$ and $x_0$ is available to the users, the prior terms in the cost function can serve as regularization terms to make the optimization process more robust.

\subsection{Initialization} \label{section: initialization}
To solve the non-convex optimization problems in the previous sections, the optimization algorithms must have good initial values for the state trajectory $x_{0:T}$ and the calibration parameter $\theta$, as well as the covariance of the noise $\Sigma_{(\cdot)}$ if fast convergence and good accuracy are desired. The covariance of the noise can be obtained from the sensor datasheet or estimated using the Allan variance analysis~\cite{Farrell2022Inertial}. To initialize the state trajectory, the initial gyroscope bias is estimated by calculating the mean value from stationary data. The sensor's stationary phase can be detected by a zero angular rate detector~\cite{Skog2010Zero}. This bias is then subtracted from the measurements, and the corrected values are integrated to obtain an initial estimate of the orientation trajectory. The initial orientation is assumed to be an identity matrix, which is approximately correct when the sensor board is held flat with its Z-axis pointing upwards and X-axis aligned with the north using a digital compass on a smartphone.
Regarding IMU bias initialization, $o^a$ is initialized to zero, and $o^\omega$ is initialized to the mean value of the stationary data. The reason the accelerometer bias $o^a$ is initialized to zero is that it cannot be estimated from stationary data unless the sensor is placed on a perfectly horizontal surface. Further, the dip angle $\alpha$ is initialized using a geomagnetic calculator based on the most recent world magnetic model~\cite{NOAA_Geomag_Calculator} and the current location. To initialize the calibration parameters $D^m$ and $o^m$, intrinsic calibration and extrinsic calibration are applied.

\subsubsection{Intrinsic Calibration} 
\begin{figure}[tb!]
    \centering
    \includegraphics{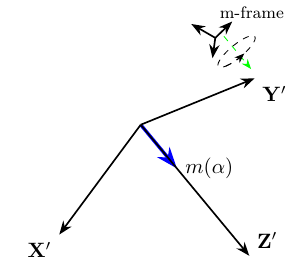}
    \caption{The reference (big axes sets) and magnetometer's coordinate frames (small axes sets) used in the intrinsic calibration. The Z-axis in the reference frame is aligned with the local magnetic field, $m$. The orientation of the m-frame can be parameterized with only roll ($\phi$) and pitch ($\gamma$) angles, since the yaw ($\psi$) angle represents a rotation around the magnetic field direction (in green), which does not change the relative inclination of the sensor to the field.}
    \label{fig: coordinate frame 2}
\end{figure}

It refers to the calibration of the sensor biases, non-orthogonal sensitivity axes, and scale factors of the magnetometers, using magnetometer measurements only. 
To this end,~\eqref{eq: factorized distortion matrix} is used to substitute $D^m$ in~\eqref{eq: magnetometer model} and $R_D$ is absorbed into a time varying orientation matrix $R^m_k=R_D R_k^\top$, resulting in the following measurement model
\begin{subequations}
    \begin{align}
    \tilde{m}_k &=  h^m(R_{k}^{m};\theta) + e_k^{m}\\
    \intertext{where}
    h^m(R_{k}^{m};\theta) &= D^I R_{k}^{m} m^{\text{n}^\prime} + o^m.
    \end{align}
\end{subequations}
Here $R_k^m$ denotes the rotation matrix that aligns the magnetometer frame with the reference frame $\text{n}^\prime$. For convenience, the reference frame is chosen such that the positive Z-axis aligns with the magnetic field vector, see Fig.~\ref{fig: coordinate frame 2}. The benefit of this particular choice is that the orientation can be fully described using only the roll ($\phi$) and pitch ($\gamma$) angles. This is because yaw ($\psi$) represents a rotation around the magnetic field direction, which does not change the relative inclination of the sensor to the field. 

Let $R_k^m$ be defined in terms of roll and pitch angles, i.e.,
\begin{subequations}
\begin{equation}
    R_k^m \triangleq R_k^m(\phi_k,\gamma_k)=R_x(\phi_k)R_y(\gamma_k)
\end{equation}
where 
\begin{align}
    R_x(\phi_k) &= \begin{bmatrix}
        1 & 0 & 0\\
        0 & \text{cos}(\phi_k) & \text{sin}(\phi_k) \\
        0 & - \text{sin}(\phi_k) & \text{cos}(\phi_k) 
    \end{bmatrix}\\
    R_y(\gamma_k) &= \begin{bmatrix}
        \text{cos}(\gamma_k) & 0 & -\text{sin}(\gamma_k)\\
        0 & 1 & 0 \\
        \text{sin}(\gamma_k) & 0 & \text{cos}(\gamma_k)
    \end{bmatrix}.
\end{align}
\end{subequations}

The ML estimate of the intrinsic parameters can be obtained by solving 
   \begin{equation}
        \label{eq: intrinsic calibration}
        \begin{split}
        &\left(\hat{D^{I}}, \hat{o}^m, \hat{\phi}_{1:T}, \hat{\gamma}_{1:T} \right)^{\text{\tiny{ML}}} \\
        &= \argmin_{D^{I}, o^m, \phi_{1:T}, \gamma_{1:T} } \sum_{k=1}^T \|\tilde{m}_k\! -\! D^{I} \!R_k^m(\phi_k,\! \theta_k) m^{\text{n}^\prime} \!- \!o^m\|^2.
        \end{split}
    \end{equation}
Here $m^{\text{n}^\prime} = (0\;0\;1)^{\top}$ is the normalized magnetic field vector expressed in the reference frame $\text{n}^\prime$. The solution to~\eqref{eq: intrinsic calibration} is used to initialize the intrinsic calibration matrix $D^I$ and the magnetometer bias $o^m$.

\subsubsection{Extrinsic Calibration} It refers to the calibration of the rotation matrix $R_D$ that aligns the magnetometer frame with the inertial sensor frame. In this work, we adopt the well-established method proposed in~\cite{Wu2016Misalignment}, which aligns the magnetometer frame with that of the gyroscope. This approach leverages the fact that, in a homogeneous magnetic field, variations in magnetometer measurements are solely caused by orientation changes, as measured by the gyroscope. Interested readers are referred to~\cite{Wu2016Misalignment} for further details.
\subsection{Summary and Practical Issues}
A summary of the proposed calibration method is shown in Algorithm~\ref{alg:calibration}. The following are some practical issues that need to be considered when applying the proposed method.
\begin{algorithm}[t]
\caption{The Proposed Calibration Method}
\label{alg:calibration}

\begin{algorithmic}
\State \textbf{1) Parameter and rotation trajectory initialization}
\Statex\hspace{\algorithmicindent}  a) Initialize $\hat{o}^a$ with zeros.
\Statex\hspace{\algorithmicindent}  b) Initialize $\hat{o}^{\omega}$ by averaging gyroscope measurements collected while the sensor is stationary.
\Statex\hspace{\algorithmicindent}  c) Initialize $\hat{x}_{0:T}$ by integrating the gyroscope measurements after the gyroscope bias in step b) has been removed.
\Statex\hspace{\algorithmicindent}  d) Initialize $\hat{\alpha}$ with local magnetic field dip angle.
\Statex\hspace{\algorithmicindent}  e) Initialize $\hat{D}^m = \hat{D}^{I^{\text{\tiny{ML}}}}\hat{R}_D^{\text{\tiny{ML}}}$ and
              $\hat{o}^m = \hat{o}^{m^{\text{\tiny{ML}}}}$ using intrinsic and extrinsic calibration in Section~\ref{section: initialization}.
\State \textbf{2) Solve the optimization problem in~\eqref{eq: original MAP} or~\eqref{eq: MAP preintegration},
       depending on whether IMU preintegration is used.}
\end{algorithmic}
\end{algorithm}

\subsubsection{Observability} Observability is a critical aspect of estimation; to accurately estimate calibration parameters, one must ensure that all states are observable. An analysis of a similar magnetometer-IMU calibration problem is provided in~\cite{yuanxin2018Dynamic}, which establishes two sufficient conditions for an observable system. These two conditions concern the initial orientation and measurements. However, it is difficult to generalize these conditions into which motion patterns render the calibration parameters observable. To gain practical insight into how observability can be ensured, the problem can instead be interpreted from a system identification perspective. According to~\cite[Ch. 13]{Lennart1999SystemIdentification}, the system input must be ``rich". In this context, this necessitates orienting the sensor system in as many directions as possible~\cite{kok2016magnetometer}. A recommended sequence is to rotate $180^{\circ}$ in yaw, pitch, and roll, respectively, and then continue rotating to maximize spatial coverage.
\subsubsection{Inhomogeneous and Non-Static Magnetic Fields} An assumption made in this work is that the calibration data can be collected in homogeneous static magnetic fields. This assumption holds when magnetometers, whose noise level is above several tens of nano Tesla, are placed outdoors free from magnetic field disturbance, e.g., an open field. However, when the calibration must be done in an environment where the magnetic field is subject to the influence of ferromagnetic materials and electromagnetic pulses, the authors recommend that users carefully rotate the sensors to minimize displacement to reduce the impact of inhomogeneous magnetic fields, and to stay away from electric equipment that generates a strong magnetic field. A good calibration performance in this case cannot be guaranteed.

\subsubsection{Synchronization} the proposed method assumes the magnetometer and IMU are synchronized, since most 9 degrees of freedom IMUs on the market have a built-in synchronization mechanism. However, if the magnetometer and IMU are not perfectly synchronized, the synchronization error can be estimated as a part of the calibration process using the method outlined in ~\cite{Skog2011TimeSync}.

\subsubsection{Convergence} A convergence assessment is crucial in iterative optimization to ensure solutions are reliable. One can check whether the optimization solver converges by monitoring the cost function, the gradient norm, and the step size. The cost function should decrease or stabilize, while a small gradient norm and step size indicate proximity to a stationary point~\cite{Anders2023Optimization}. Moreover, one can evaluate the quality of the optimized calibration parameter by performing cross-validation~\cite[p. 32]{bishop2006pattern} and residual analysis~\cite{martin2017fitting}.

\section{Evaluation}
The calibration methods from~\cite{kok2016magnetometer} and~\cite{yuanxin2018Dynamic} were implemented in this work. For clarity, we refer to them as the Kok et al.'s method and the Wu et al.'s method, respectively, throughout the rest of the paper. The optimization toolbox Manopt~\cite{manopt} was used to implement both the proposed method and the method by Kok et al.. We evaluated all methods on both simulated and real-world datasets. To ensure fair comparisons, identical termination criteria were applied to the optimization procedures for the proposed method and those in Kok et al.'s method. Specifically, the termination condition was set to either a norm of the update step less than $10^{-6}$ or reaching the maximum number of iterations, 400.

Regarding performance, the evaluation on the simulated datasets is based on the computation time of each method and root mean square error~(RMSE) of the estimated calibration parameters, while the evaluation on the real-world datasets is based on the computation time of each method and error-to-traveled-distance ratio\footnote{The error-to-traveled-distance ratio is the ratio between the final positioning error and the total traveled distance.} of the navigation system in~\cite{huang2023mains} using the calibrated IMU and magnetometer data.

\subsection{Simulation Study}
To evaluate the performance of all three methods under varying conditions, two sets of experiments were conducted:
\begin{enumerate}
    \item \textbf{Sampling Frequency Variation:}  
    Datasets were generated with different sampling frequencies to assess how computational efficiency and calibration accuracy scale with measurement rate.

    \item \textbf{Sampling Rate Ratio Variation:}  
    In this experiment, datasets were created with different sample rate ratios between the IMU and the magnetometer, simulating asynchronous sensor setups commonly encountered in practice.
\end{enumerate}
For each sampling frequency and sample rate ratio, 10 independent datasets were generated. In each dataset, the sensors were rotated around 6 fixed (with minor deviation) rotation axes, with an angular velocity of approximately 7 deg/s, consecutively. The duration of each dataset is approximately 5 minutes. The sensor data was generated using the sensor models in~\eqref{eq: accelerometer model} and~\eqref{eq: magnetometer model}. Further, the calibration parameters were drawn from Gaussian distributions. In particular, the distortion matrix was constructed as 
\begin{subequations}
    \begin{align}
        D^m &= D_{\text{diag}} D_{\text{skew}} R_D   \\
    \intertext{where}
    D_{\text{diag}} &\triangleq \begin{bmatrix}
        D_{1,1} & 0 & 0\\
        0 & D_{2,2} & 0\\
        0 & 0 & D_{3,3}
    \end{bmatrix} \\
    \intertext{and}
    D_{\text{skew}} &\triangleq \begin{bmatrix}
        1 & 0 & 0\\
        \text{sin}(\zeta) & \text{cos}(\zeta) & 0\\
        -\text{sin}(\eta) & \text{cos}(\eta) \text{sin}(\rho) & \text{cos}(\eta) \text{cos}(\rho)
        \end{bmatrix}.
\end{align}
\end{subequations}
Here, $D_{1,1}, D_{2,2}, D_{3,3}$ denote the different scale factors on the three sensitivity axes, respectively. Further, $\zeta, \eta, \rho$ denote the different non-orthogonality angles, respectively. Lastly, $R_D$ is defined by a set of Euler angles $\left(\phi_D, \gamma_D, \psi_D \right)$.
The calibration parameter and noise settings are summarized in the TABLE~\ref{T: calib settings}. 

\begin{figure}[tb!]
    \centering

    \subfloat[]{%
        \includegraphics[width=\columnwidth]{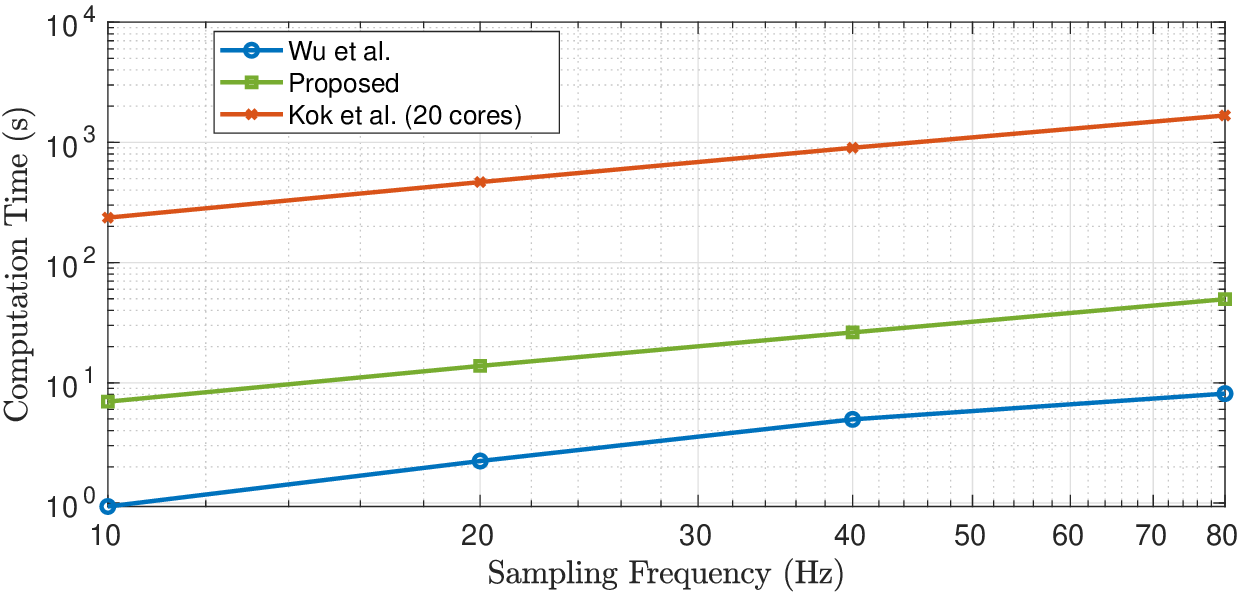}%
        \label{fig: comparison of computation time (simulation)}%
    }\\[0.5em]

    \subfloat[]{%
        \includegraphics[width=0.47\columnwidth]{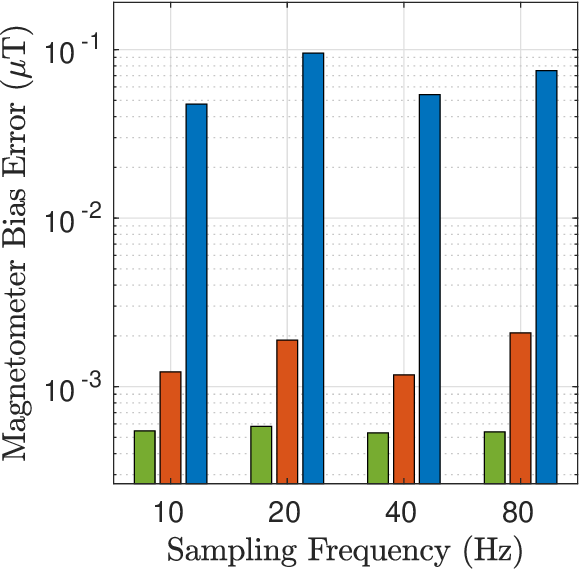}%
        \label{fig: comparison of magnetometer bias error (simulation)}%
    }\hfill
    \subfloat[]{%
        \includegraphics[width=0.47\columnwidth]{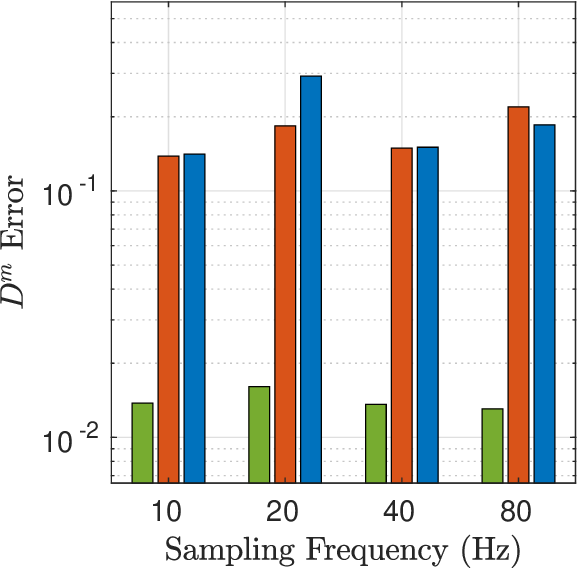}%
        \label{fig: comparison of D error (simulation)}%
    }\\[0.5em]

    \subfloat[]{%
        \includegraphics[width=0.47\columnwidth]{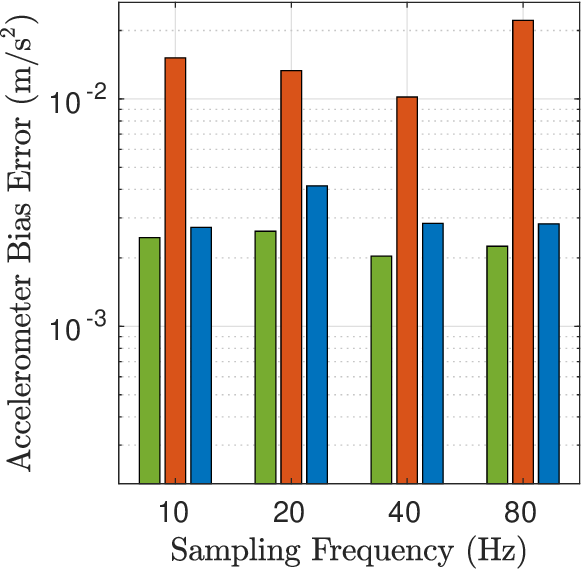}%
        \label{fig: comparison of ACC bias error (simulation)}%
    }\hfill
    \subfloat[]{%
        \includegraphics[width=0.47\columnwidth]{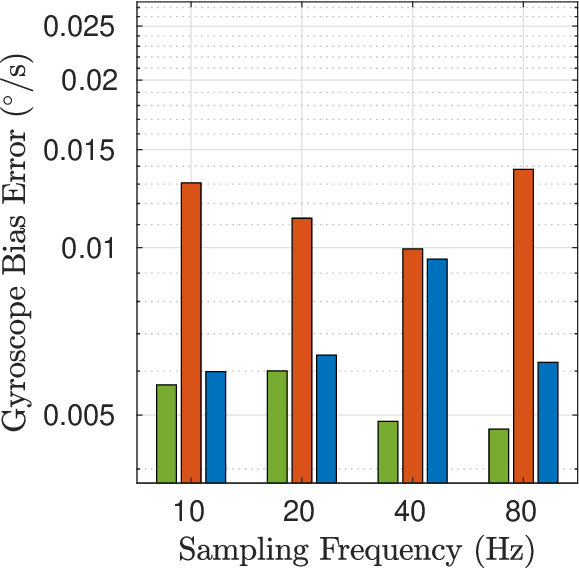}%
        \label{fig: comparison of Gyroscope bias error (simulation)}%
    }

    \caption{Comparison of the computation time (a) and RMSE of the estimated calibration parameters (b - e) on datasets with different sampling frequencies. }
    \label{fig: comparison of computation time and RMSE (simulation)}
\end{figure}

First, the evaluation was performed on the datasets with different sampling frequencies, and the results are shown in Fig.~\ref{fig: comparison of computation time and RMSE (simulation)}. The maximum sampling frequency is set to 80 Hz because it is sufficient to capture the ``slow rotation" during calibration data collection. It can be seen from Fig.~\ref{fig: comparison of computation time (simulation)} that the computation time of all three methods increases almost linearly with the sampling frequency, which implies that the processing time increases linearly with the number of sensor measurements. This is consistent with the computational complexity upper bounds discussed in Section \ref{discussion}.
Furthermore, the method by Wu et al. is the fastest, followed by the proposed method and the method by Kok et al.. Note that parallel computing with 20 CPU cores is used in computing the numerical Jacobian matrix for the method by Kok et al.; computing with a single CPU core takes a much longer time. As shown in~\cref{fig: comparison of magnetometer bias error (simulation)} and~\cref{fig: comparison of D error (simulation)}, the proposed method achieves the lowest RMSE for both bias and distortion matrix estimation. Meanwhile, the method by Kok et al. has a lower RMSE in bias than the method by Wu et al., while they share similar performance in distortion matrix estimation. In terms of accelerometer and gyroscope calibration results shown in~\cref{fig: comparison of ACC bias error (simulation)} and~\cref{fig: comparison of Gyroscope bias error (simulation)}, the proposed method still has the lowest RMSE. However, the method by Wu et al. performs better than the method by Kok et al. Another interesting observation is that the RMSEs of the calibration parameters for all methods remain approximately constant as the sampling frequency increases. A possible explanation is that the sensor's motion, specifically its rotational dynamics, has limited bandwidth due to the relatively slow movement, making lower sampling frequencies sufficient to capture the necessary information.

\begin{table}[tb]
    \centering
    \begin{threeparttable}
        \caption{Calibration parameters and noise settings used in the simulations}
        \label{T: calib settings}
        \begin{tabular}{l l}
            \toprule 
            Calibration Parameter & Distribution \\
            \midrule 
            $D_{\text{diag}}$ & $D_{1,1}, D_{2,2}, D_{3,3} \sim \mathcal{U}(0.9,1.1)$ \\
            $D_{\text{skew}}$ & $\zeta, \eta, \rho \sim \mathcal{U}(-10^{\circ},10^{\circ})$ \\
            $R_D$       &  $\phi_D, \gamma_D, \psi_D  \sim \mathcal{U}(-5^{\circ},5^{\circ})$ \\
            $o^a$ & $o^a_1, o^a_2, o^a_3 \sim  \mathcal{U}(-0.5,0.5)$ \quad (m/s$^2$) \\
            $o^\omega$ & $o^\omega_1, o^\omega_2, o^\omega_3 \sim  \mathcal{U}(0.47,0.67)$ \quad ($^{\circ}$/s) \\
            $o^m$ & $o^m_1, o^m_2, o^m_3 \sim  \mathcal{U}(-2,2)$ \quad ($\mu$T) \\
            $\alpha$ & $\mathcal{U}(67^\circ, 77^\circ)$ \\
            \midrule 
            Sensor Type & Noise Density (per $\sqrt{\text{Hz}}$) \\
            \midrule
            Accelerometer & 0.02 m/s$^2$  \\
            Gyroscope  &  0.05 $^{\circ}$/s \\ 
            Magnetometer & 0.003 $\mu$T \\ 
            \bottomrule
        \end{tabular}
        \begin{tablenotes}
            \item[1] $\mathcal{U}(a,b)$ denotes uniform distribution on $(a,b)$.
        \end{tablenotes}
    \end{threeparttable}
\end{table}

\begin{figure}[tb!]
    \centering

    \subfloat[]{%
        \includegraphics[width=\columnwidth]{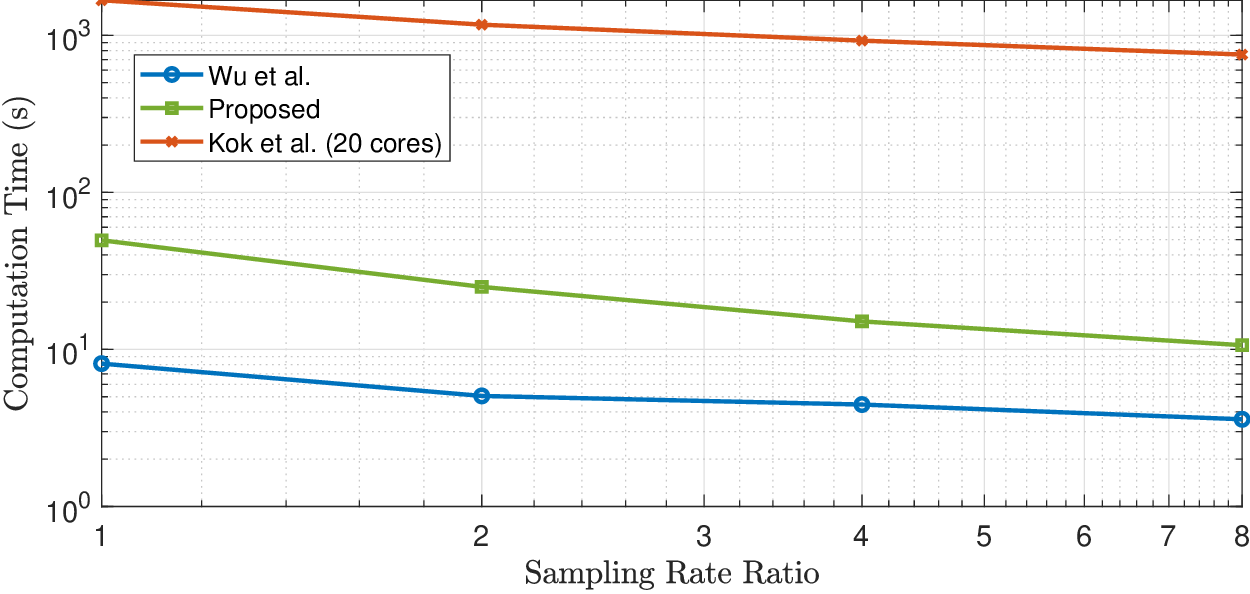}%
        \label{fig: comparison of computation time Ratio (simulation)}%
    }\\[0.5em]

    \subfloat[]{%
        \includegraphics[width=0.47\columnwidth]{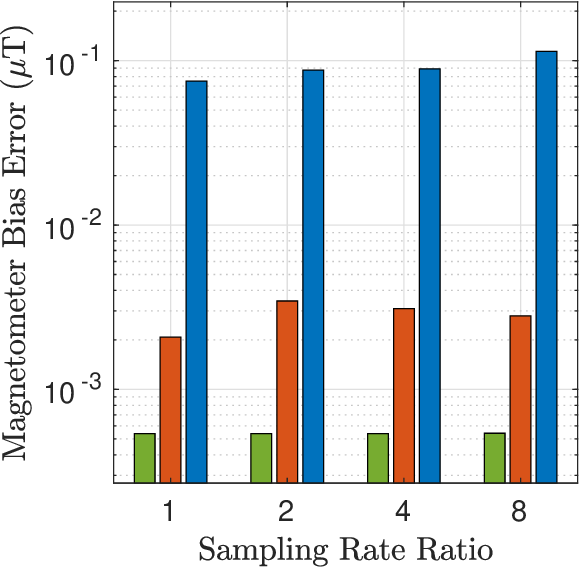}%
        \label{fig: comparison of magnetometer bias error Ratio (simulation)}%
    }\hfill
    \subfloat[]{%
        \includegraphics[width=0.47\columnwidth]{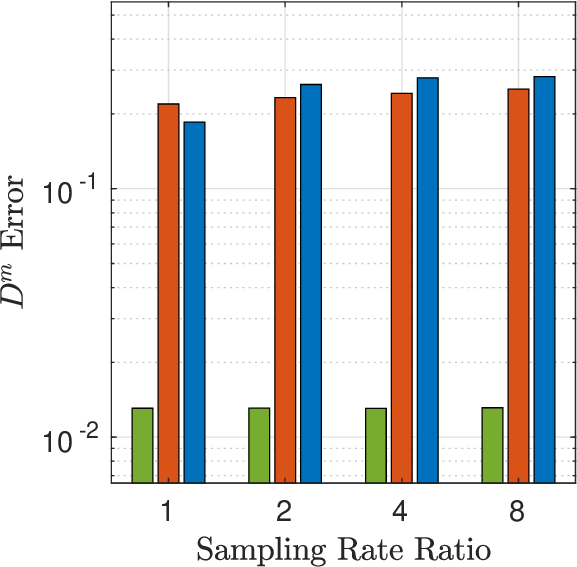}%
        \label{fig: comparison of D error Ratio (simulation)}%
    }\\[0.5em]

    \subfloat[]{%
        \includegraphics[width=0.47\columnwidth]{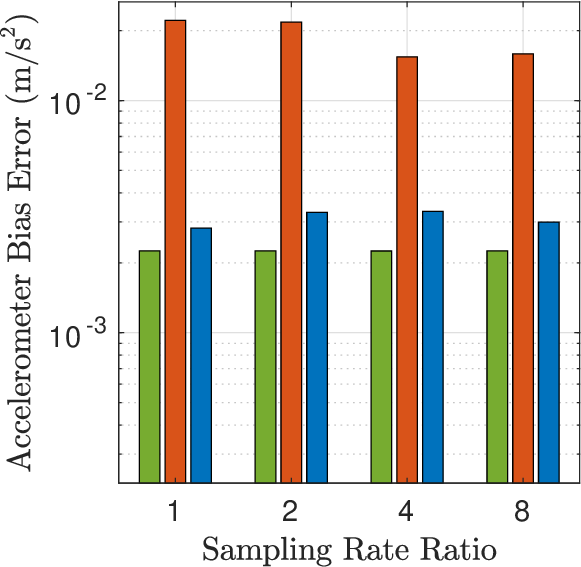}%
        \label{fig: comparison of ACC bias error Ratio (simulation)}%
    }\hfill
    \subfloat[]{%
        \includegraphics[width=0.47\columnwidth]{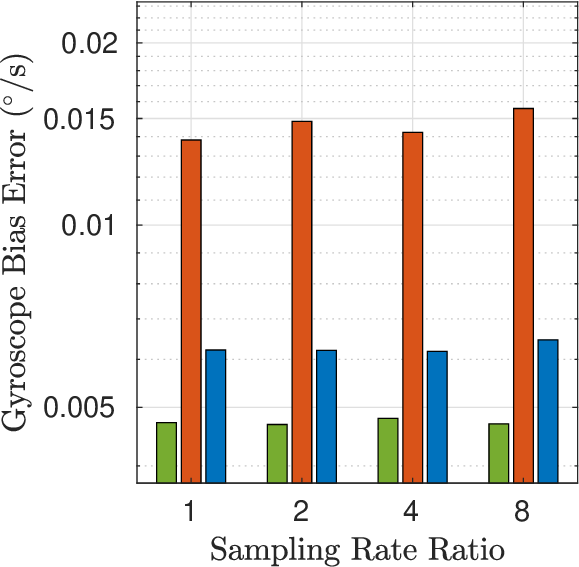}%
        \label{fig: comparison of Gyroscope bias error Ratio(simulation)}%
    }

    \caption{Comparison of the computation time (a) and RMSE of the estimated calibration parameters (b - e) on datasets with different sampling rate ratios. The horizontal axis is the ratio of the IMU sampling rate to the magnetometer sampling rate.}
    \label{fig: comparison of computation time and RMSE Ratio (simulation)}
\end{figure}

Next, the three methods were evaluated on datasets where the magnetometer sampling rate was several times slower (specifically 2, 4, and 8 times) than that of the IMU (80 Hz). These ratios define the sampling rate ratio, which refers to the ratio between the IMU and magnetometer sampling rates. For the proposed method, the cost function was constructed according to~\eqref{eq: MAP preintegration}. 
The other two methods performed EKF updates only when magnetometer and accelerometer measurements were received simultaneously. In this way, all three methods process the same number of measurements.
The results are shown in Fig.~\ref{fig: comparison of computation time and RMSE Ratio (simulation)}. It can be seen that the computation time of all three methods decreases as the sampling rate ratio increases, while the RMSE of the estimated calibration parameters remains approximately constant. This indicates that one can have a lower magnetometer sampling rate than the IMU sampling rate to save computation time without sacrificing the accuracy of the calibration parameters. Moreover, \cref{fig: comparison of D error Ratio (simulation),fig: comparison of ACC bias error Ratio (simulation),fig: comparison of magnetometer bias error Ratio (simulation),fig: comparison of Gyroscope bias error Ratio(simulation)} show that the relative RMSE magnitudes ranking across all three methods remains consistent as in the previous experiment.

To evaluate the proposed method's sensitivity to the initial value of the gyroscope bias, the proposed algorithm is initialized with different bias values perturbed around the true value by varying magnitudes, from $0$ to $10^{-1}$~rad/s. The resulting RMSEs are the same as when the initialization error is less than or equal to $10^{-3}$~rad/s, but the optimization program failed to converge when the initialization error increased to $10^{-2}$ rad/s. However, an initialization error of this magnitude is rare after the initialization step described in~\ref{section: initialization}. It can be concluded that the proposed algorithm is capable of correcting the initialization error of the gyroscope bias to some extent.

Lastly, the RMSE of the estimated calibration parameters under different synchronization errors is reported in TABLE~\ref{Tab: RMSE wrt Sync. Err.}. The results show that the RMSEs become larger when there is a synchronization error between the magnetometer and IMU. The most sensitive parameters are accelerometer bias and $D^m$. Nevertheless, the RMSE errors of the estimated calibration parameters remain small. 

\begin{table}[t]
\centering
\begin{threeparttable}
\caption{RMSE of the Estimated Calibration Parameters under Different Synchronization Errors (Sample Rate: 80 Hz)}
\begin{tabular}{lccc}
\toprule
\diagbox{RMSE}{Error [ms]} & -12.5  & 0  & 12.5 \\
\midrule
Accelerometer bias (m/s$^2$) & 0.0031 & 0.0022 & 0.0040 \\
Gyroscope bias (rad/s)       & 8.9$\times$10$^{-5}$ & 8.2$\times$10$^{-5}$ & 9.1$\times$10$^{-5}$ \\
Magnetometer bias ($\mu$T)   & 0.0005 & 0.0005 & 0.0005 \\
$D^m$                        & 0.0330  & 0.0130  & 0.0269 \\
\bottomrule
\end{tabular}
\begin{tablenotes}
\item The synchronization errors refer to the delays of the magnetometer. 
\end{tablenotes}
\label{Tab: RMSE wrt Sync. Err.}

\end{threeparttable}
\end{table}

\subsection{Real-world Datasets Study}
\begin{figure}[tb!]
    \centering
    \includegraphics[width=\columnwidth]{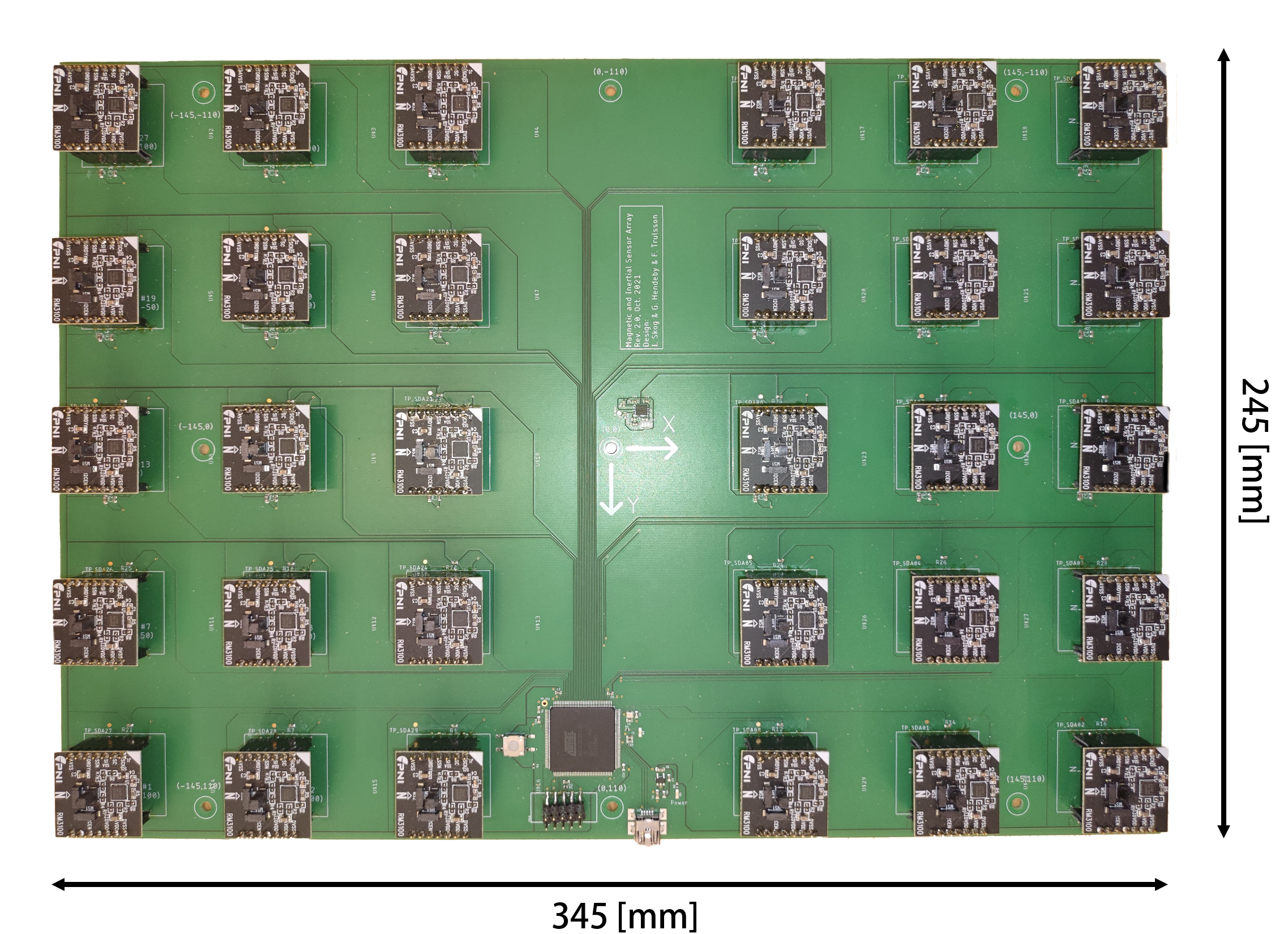}
    \caption{The sensor board used in the experiment. It has 30 PNI \href{https://www.pnicorp.com/rm3100/}{RM3100} magnetometers and an \href{https://www.inertialelements.com/osmium-mimu4444.html}{Osmium MIMU 4844 IMU} mounted on the bottom side.}
    \label{F: sensor board}
\end{figure}

In real-world experiments, the three methods were used to calibrate a magnetic and inertial sensor array (see Fig.~\ref{F: sensor board}), which consists of 1 IMU and 30 magnetometers. The IMU and magnetometers are low-cost, off-the-shelf sensors used for navigation, and their main characteristics are summarized in TABLE~\ref{T: specifications}. The inductive coupling effect caused by nearby magnetometers was neglected due to the limited measurements of other physical quantities.

\begin{table}[tb!]
\centering
\caption{IMU and Magnetometer Specifications}
\label{tab:imu_magnetometer_specs}
\begin{threeparttable}
\begin{tabular}{lcc}
\toprule
\textbf{Parameter} & \textbf{IMU} & \textbf{Magnetometer} \\
\midrule
Model & Osmium MIMU 4844 & PNI
RM3100 \\
Price & \$799&\$25\\
Sensitivity& --- & 13 nT\\
Noise Density & --- & 1.2 nT / $\sqrt{\text{Hz}}$\\
Bias Stability &  \makecell{Accelerometer: 0.03 mg \\Gyroscope: 0.07$^{\circ}$ / h}  & ---\\
Sample Rate & 62.5 Hz& 62.5 Hz\\
\bottomrule
\end{tabular}
\begin{tablenotes}
\item ---: not provided in the manual.
\end{tablenotes}
\end{threeparttable}
\label{T: specifications}
\end{table}

\begin{figure}[tb!]
    \centering

    \subfloat[]{%
        \includegraphics[width=0.98\columnwidth]{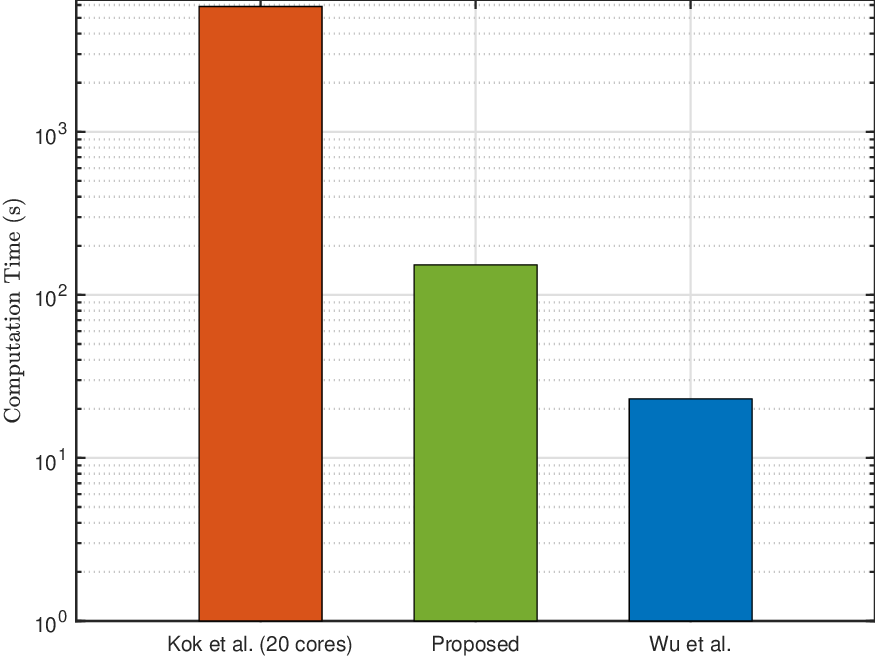}%
        \label{F: real computation time}%
    }\\[0.2em]

    \subfloat[]{%
        \includegraphics[width=0.98\columnwidth]{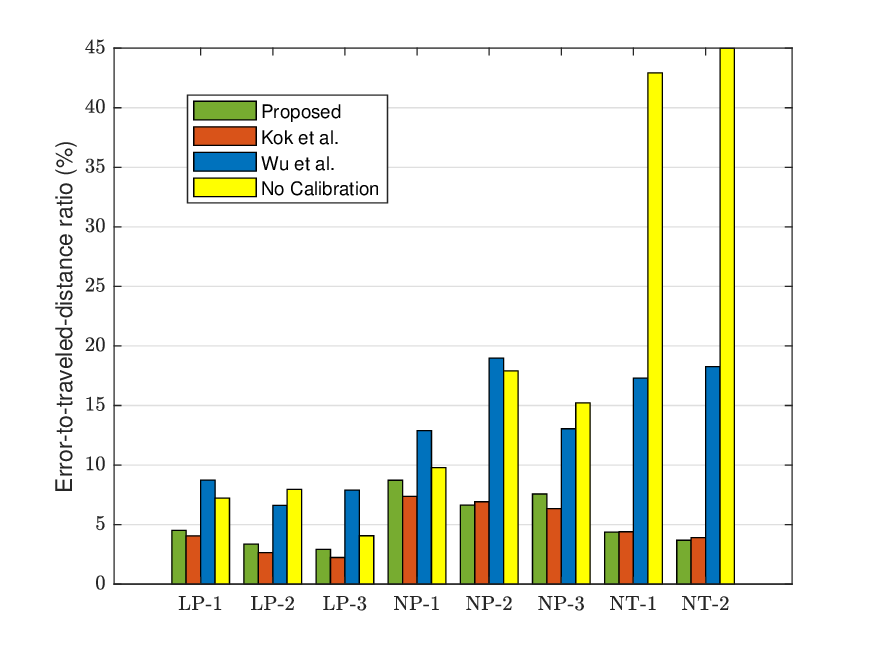}%
        \label{F: real position error}%
    }\\[0.2em]

    \subfloat[]{%
        \includegraphics[width=0.98\columnwidth]{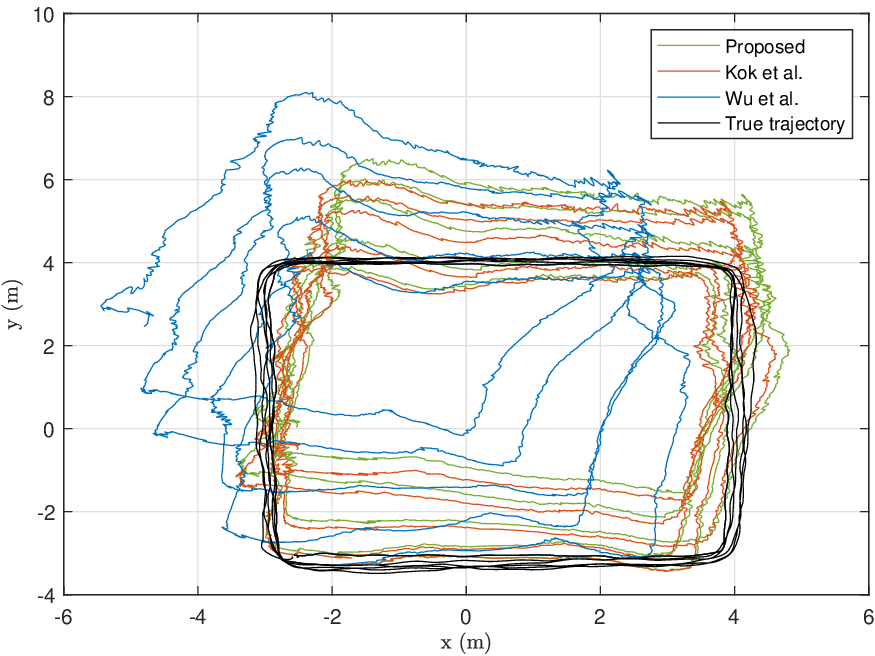}%
        \label{F: trajetory}%
    }

    \caption{Performance comparison on real-world data. (a) The computation time of the calibration algorithms. (b) The error-to-traveled-distance ratios of a magnetic field-aided inertial navigation system that uses the uncalibrated and calibrated magnetometer data. The labels on the x-axis represent different datasets. (c) The horizontal trajectories (LP-2) estimated by the magnetic field-aided inertial navigation system with calibrated data vs. the true trajectory.}
    \label{F: real position error and computation time}
\end{figure}

The calibration data was collected on June 1$^{\text{st}}$, 2024 in Linköping, Sweden, on an outdoor grass field, where the homogenous magnetic field assumption can be considered valid given the size of the array and the sensitivity of the sensors~\cite{kok2016magnetometer}. The magnetometer and IMU's sample rate was 62.5~Hz, with data synchronized within 16 ms. The duration of the calibration data collection was approximately 5 minutes. During the data collection, the sensor array was rotated slowly by hand so that the assumption of zero acceleration is reasonably valid, and the sensor platform was placed into as many orientations as possible. The magnetometer and accelerometer data were downsampled by a factor of 3 to save computation cost. The joint magnetometer-IMU calibration procedures were done for each magnetometer-IMU pair. To evaluate the performance of the three methods, the estimated calibration parameters were used to calibrate the magnetometer data used in~\cite{huang2023mains}, and the calibrated data were fed into the magnetic field-aided inertial navigation system (MAINS) proposed in~\cite{huang2023mains}. The computation time of the three calibration methods, the error-to-traveled-distance ratio of the MAINS, and an example of output trajectories are shown in Fig.~\ref{F: real position error and computation time} (a) - (c), respectively. As a first observation, the ranking of the methods by calibration computation time matches the simulation results. Second, the joint magnetometer-IMU calibration is clearly essential for MAINS to achieve good performance, since uncalibrated data leads to significantly higher error-to-traveled-distance ratios as shown in Fig.~\ref{F: real position error}. The proposed method and Kok et al.'s method perform better than Wu et al.'s method in terms of the error-to-traveled-distance ratio, which is consistent with the simulation results. Although the proposed method results in a slightly higher error-to-traveled-distance ratio than that of Kok et al., it significantly outperforms the latter in terms of computational efficiency, being approximately one order of magnitude faster. This slight reduction in accuracy may be due to Kok et al.'s method being more robust to unmodeled sensor errors. The computational efficiency of the Kok et al.'s method is limited by the use of numerical derivatives instead of analytic ones, due to the difficulty of deriving closed-form Jacobians, as discussed in Section~\ref{discussion}.

To evaluate our proposed method's sensitivity to IMU sensor quality, we simulated an automotive-grade IMU in calibration by utilizing data from only a single IMU sensor out of the 32-IMU array provided by the \href{https://www.inertialelements.com/osmium-mimu4444.html}{Osmium MIMU4444} device\footnote{In the previous calibration experiment, we used the averaged output of the 32-IMU array provided by the Osmium MIMU4444.}. After obtaining the calibration parameters, we calibrated the magnetometer data used in the positioning experiment, then repeated the experiment, and compared the final positioning errors with those from the previous experiment. To isolate the effect of the IMU quality, we used the same IMU data from the previous positioning experiment. The final positioning errors for both experiments are shown in TABLE~\ref{T: RMSE comparison}. The results show that both sets of calibration parameters yield similar positioning errors across all datasets, demonstrating the proposed method's robustness and its capability to jointly calibrate a magnetometer alongside an automotive-grade IMU.

\begin{table}[tb!]
\centering
\caption{The Comparison of Final Positioning Errors Resulting From the Two Sets of Calibration Parameters $\theta$ (Unit: Meter)}
\begin{threeparttable}
\begin{tabular}{lcc}
\toprule
\textbf{Dataset} & $\theta$ (High-end IMU) & $\theta$ (Low-end IMU) \\
\midrule
LP-1 & 5.14 & 5.17 \\
LP-2 & 4.69& 4.72\\
LP-3& 4.73 & 4.72\\
NP-1& 8.14 & 7.95\\
NP-2& 5.90 & 6.78\\
NP-3& 6.50 & 6.81\\
NT-1& 5.33& 5.72\\
NT-2& 3.19& 3.37\\
\bottomrule
\end{tabular}
\end{threeparttable}
\label{T: RMSE comparison}
\end{table}

\section{Conclusion}
In this paper, we propose a new method for joint magnetometer-IMU calibration, which jointly estimates the calibration parameters and orientation trajectory. The proposed method was compared with two state-of-the-art methods in terms of computational efficiency and calibration accuracy. Simulation results showed that the proposed method achieves lower RMSE in calibration parameters compared to the other methods, reducing accuracy by 20–30\%, while maintaining competitive computational efficiency. Real-world experiments demonstrated that the proposed method can effectively calibrate an inertial-magnetometer array, thereby reducing position drift in a magnetic field-aided inertial navigation system. The proposed method is one order of magnitude faster than the most accurate state-of-the-art algorithm implemented in this work, while delivering comparable accuracy. These results suggest that the proposed method is an attractive alternative to existing state-of-the-art methods, offering fast and accurate calibration.

\appendices{
\section{Gauss Newton Method on Manifolds} \label{app:GN}
A general nonlinear least squares problem may be expressed as
\begin{equation}
    \hat{x} = \argmin_{x} \sum_{k} \|g_k(x)\|^2 \label{eq: nonlinear least squares}
\end{equation}
where $x \in \mathcal{M}$, $\mathcal{M}$ is a manifold, and $g_k(x)$ is a vector-valued function. 
To solve~\eqref{eq: nonlinear least squares} with the Gauss-Newton method, one must first specify a operation that allows the Jacobian of $g_k(x)$ to be computed as well as the increment $\delta x \in \mathbb{R}^{\text{dim}(\mathcal{M})}$ to be applied to $x$ to obtain a new point on the manifold. 

In \eqref{eq: original MAP} and \eqref{eq: MAP preintegration}, $x \in \mathcal{M}= \mathbb{R}^{\text{dim}(\theta)} \times \SO^M$, where $M$ is the number of rotation matrices. The operation associated with the manifold $\mathcal{M}$ is given by
\begin{subequations}\label{eq: retraction}
    \begin{align} 
        \theta \oplus \delta \theta &\triangleq \theta + \delta \theta \\
        R_i \oplus \delta R_i & \triangleq \text{Exp}(\delta R_i)R_i .
    \end{align}
\end{subequations}
Here $\delta R_i \in \mathbb{R}^3$ is the increment in the tangent space of $\SO$. Furthermore, the Jacobian of $g_k(x)$ is given by
\begin{equation}
J_k(x) = \left[ 
    \frac{\partial g_k}{\partial \theta} \;\;
    \frac{\partial g_k}{\partial R_0} \;\;
    \cdots \;\;
    \frac{\partial g_k}{\partial R_{M-1}} 
\right]
\end{equation}
where 
\begin{align}
    \frac{\partial g_k}{\partial \theta} &\triangleq \frac{\partial g_k\big(\theta \oplus \delta \theta, \{R_i\}_{i=0}^{M-1}\big)}{\partial \delta \theta} \Bigg|_{\delta \theta=0} \\
    \frac{\partial g_k}{\partial R_i} &\triangleq \frac{\partial g_k \big(\theta, \cdots,R_i \oplus \delta R_i,\cdots \big)}{\partial \delta R_i} \Bigg|_{\delta R_i=0}.
\end{align}
With these definitions, the set of linear equations (normal equations) in each iteration of the Gauss-Newton algorithm can be constructed, and the increment obtained from solving it can be used to update the current estimate using~\eqref{eq: retraction}. 

\section{Derivatives for~\eqref{eq: original MAP}} \label{app:first}
For simplicity, the priors are assumed to be uniform, i.e., $\ln(p(\theta)), \ln(p(x_0))$ are constants; therefore, the corresponding terms can be removed from the cost function. The derivatives required by the optimization program are the derivatives of each individual residual term with respect to the calibration parameter $\theta$ and state $R_k$. Let the residual terms be defined as
\begin{align}
    r_k^a &\triangleq L_a^{\top} (\tilde{s}_k + R_k^{\top} \gravity - o^a)\\
    r^m_k &\triangleq L_m^{\top} \left( \tilde{m}_k -D^m R_k^{\top} m(\alpha) - o^m \right) \\
    r_k^\omega &\triangleq L_\omega^{\top} \left( \tilde{\omega}_k - \frac{1}{\dT} \text{Log}(R_k^{\top} R_{k+1} ) - o^\omega \right)
\end{align}
where $L_{(\cdot)}$ is the lower triangular matrix from Cholesky decomposition of the covariance matrix $\Sigma_{(\cdot)}^{-1}$.

The derivatives of $r_k^a$ with respect to $R_k$ and $o^a$ are given by
\begin{subequations}
\begin{align}
    \frac{\partial r_k^a}{\partial  R_k} &= L_a^{\top} [R_k^{\top} \gravity]^{\wedge} R_k^{\top}\\
    \frac{\partial r_k^a}{\partial  o^a} &= -L_a^{\top}
\end{align}
\end{subequations}
respectively.
The derivatives of $r_k^m$ with respect to $R_k$, $D^m$, $\alpha$ and $o^m$ are given by
\begin{subequations}
\begin{align}
    \frac{\partial r_k^m}{\partial  R_k} &= -L_m^{\top} D^m[R_k^{\top} m(\alpha)]^{\wedge} R_k^{\top}\\
    \frac{\partial r_k^m}{\partial  D^m_{\text{vec}}} &= -L_m^{\top} \left(R_k^{\top} m(\alpha) \right)^\top \otimes_K I_3\\
    \frac{\partial r_k^m}{\partial \alpha} &= L_m^{\top} D^m R_k^{\top} \big( 0\; \text{sin}(\alpha) \; \text{cos}(\alpha)\big)^{\top}\\
      \frac{\partial r_k^m}{\partial  o^m} &= -L_m^{\top}
\end{align}
\end{subequations}
respectively.
Here $D^m_{\text{vec}} \in \mathbb{R}^9$ denotes the vectorized form of $D^m$ and $\otimes_K$ denotes the Kronecker product.
The derivatives of $r_k^{\omega}$ with respect to $R_k$, $R_{k+1}$, and $o^{\omega}$ are given by
\begin{subequations}
\begin{align}
    \frac{\partial r_k^\omega}{\partial  R_k} &= \frac{1}{\dT}L_\omega^{\top}  J^{-r}\left(\text{Log}(R_k^{\top} R_{k+1})\right) R_{k+1}^{\top} \\
    \frac{\partial r_k^\omega}{\partial R_{k+1}} &= -\frac{1}{\dT}L_\omega^{\top}  J^{-r}\left(\text{Log}(R_k^{\top} R_{k+1})\right)  R_{k+1}^{\top} \\
    \frac{\partial r_k^\omega}{\partial  o^{\omega}} &= -L_\omega^{\top}
\end{align}
\end{subequations}
respectively. Here $J^{-r}(\cdot)\triangleq \big(J^r(\cdot)\big)^{-1}$.

Interested readers can derive the above expressions with~\cite{sola2018micro}, where they are referred to as \textit{Left Jacobians on Lie groups}~\cite[p.8, eq.44]{sola2018micro}.

\section{Derivatives for~\eqref{eq: MAP preintegration}} \label{app:second}
\newcommand{\dydo}{\evalat[\bigg]{\frac{\partial{ \Delta \tilde{R}_{k', k'+N}}}{\partial o^{\omega}}}{o^\omega=\bar{o}^{\omega}}}

As in Appendix B, we assume uniform priors, allowing us to omit the corresponding terms from the cost function. The derivatives of the accelerometer and magnetometer terms are the same as in Appendix B. To simplify the derivatives of the IMU preintegration term, we follow the assumption in \cite{forster2016manifold}:\\
 the nominal gyroscope bias $\bar{o}^\omega \in \mathbb{R}^3$ is treated as known. In practice, the nominal gyroscope bias can be obtained from averaging stationary data. The true gyroscope bias is then modeled as $o^\omega = \bar{o}^\omega + \Delta o^\omega$, where $\|\Delta o^\omega\|$ is small. 
This assumption avoids having to directly recompute the preintegration measurements $\Delta \tilde{R}_{i,j}$ and their covariances $\Sigma_{\Delta \tilde{R}_{i,j}}$ whenever the bias estimate changes during optimization.

With this approach, the preintegration measurements are linearized at $\bar{o}^\omega$ and approximated in each iteration as
 \begin{equation}
     \Delta \tilde{R}_{k',k'+N} \approx  \Delta \bar{R}_{k',k'+N} \text{Exp}\left(\dydo \Delta o^{\omega}\right)
\end{equation}
where $ \Delta \bar{R}_{k',k'+N} \triangleq \evalat{\Delta \tilde{R}_{k',k'+N}}{o^\omega=\bar{o}^{\omega}}$. According to~\cite{forster2016manifold}, the covariance of the preintegration measurements is
\begin{equation}
\Sigma_{\Delta \tilde{R}_{k',k'+N}} =  \sum_{k=k'}^{k'+N-1} \tilde{A}_k \Sigma_\omega \tilde{A}_k^{\top},
\end{equation}
where $ \tilde{A}_k = \Delta \tilde{R}_{k+1, k'+N - 1}^{\top} J_k^r \Delta T$. In practice, the covariance matrix is often approximated as 
\begin{equation}
\Sigma_{\Delta \tilde{R}_{k',k'+N}} \approx  \sum_{k=k'}^{k'+N-1} \bar{A}_k \Sigma_\omega \bar{A}_k^{\top},
\end{equation}
where $ \bar{A}_k = \Delta \bar{R}_{k+1, k'+N - 1}^{\top} J_k^r \Delta T$, unless $\Delta o^{\omega}$ becomes sufficiently large, in which case the covariances are re-evaluated. Next, we present the derivatives when the approximation holds.

Let the preintegration residual term be defined as
\begin{equation}
    r_{k'}^{p.} = L_{p.}^{\top} \text{Log}(\underbrace{\Delta \tilde{R}_{k', k'+N}^{\top} R_{k'}^{\top} R_{k'+N}}_{\triangleq Q} )
\end{equation}
where $L_{p.}$ is the lower triangular matrix from Cholesky decomposing $\Sigma_{\Delta \tilde{R}_{k',k'+N}}^{-1}$. 

The derivatives of $r_{k'}^{p.}$ with respect to $R_k'$, $R_{k'+N}$, and $\Delta o^{\omega}$ are given by
\begin{subequations}
    \begin{align}
        \frac{\partial r_{k^\prime}^{\text{p.}}}{\partial  R_{k^\prime}} &= - L_{\text{p.}}^{\top} J^{-r}\big(\text{Log}(Q)\big)  R_{k^{\prime}+ N}^{\top} \\
        \frac{\partial r_{k^\prime }^{\text{p.}}}{\partial  R_{k^\prime + N}} &=  L_{\text{p.}}^{\top}  J^{-r}\big(\text{Log}(Q)\big)  R_{k^{\prime}+ N}^{\top} \\
        \frac{\partial r_{k^\prime }^{\text{p.}}}{\partial  \Delta o^{\omega}}
        &= -L_{\text{p.}}^{\top}  J^{-r}\big(\text{Log}(Q)\big)     Q^{\top} \\ \nonumber
        &\quad\times J^{r}\left(\dydo \Delta o^{\omega}\right) \\ \nonumber
        &\quad \times\dydo
    \end{align}
    where 
    \begin{equation}
        \dydo= -\sum_{i=k'}^{k'+N-1} \Delta \tilde{R}_{i+1, k'+N}^{\top} J_i^r \Delta T.
    \end{equation}
\end{subequations}
}

\section*{Acknowledgments}
The authors would like to thank Manon Kok (Delft University of Technology) for inspiring the development of the ideas presented in this work, and Yuanxin Wu (Shanghai Jiao Tong University) for valuable discussions on the implementation details of~\cite{yuanxin2018Dynamic}.

\bibliographystyle{IEEEtran}
\bibliography{ref_short}

\begin{IEEEbiography}[{\includegraphics[width=1in,height=1.25in,clip,keepaspectratio]{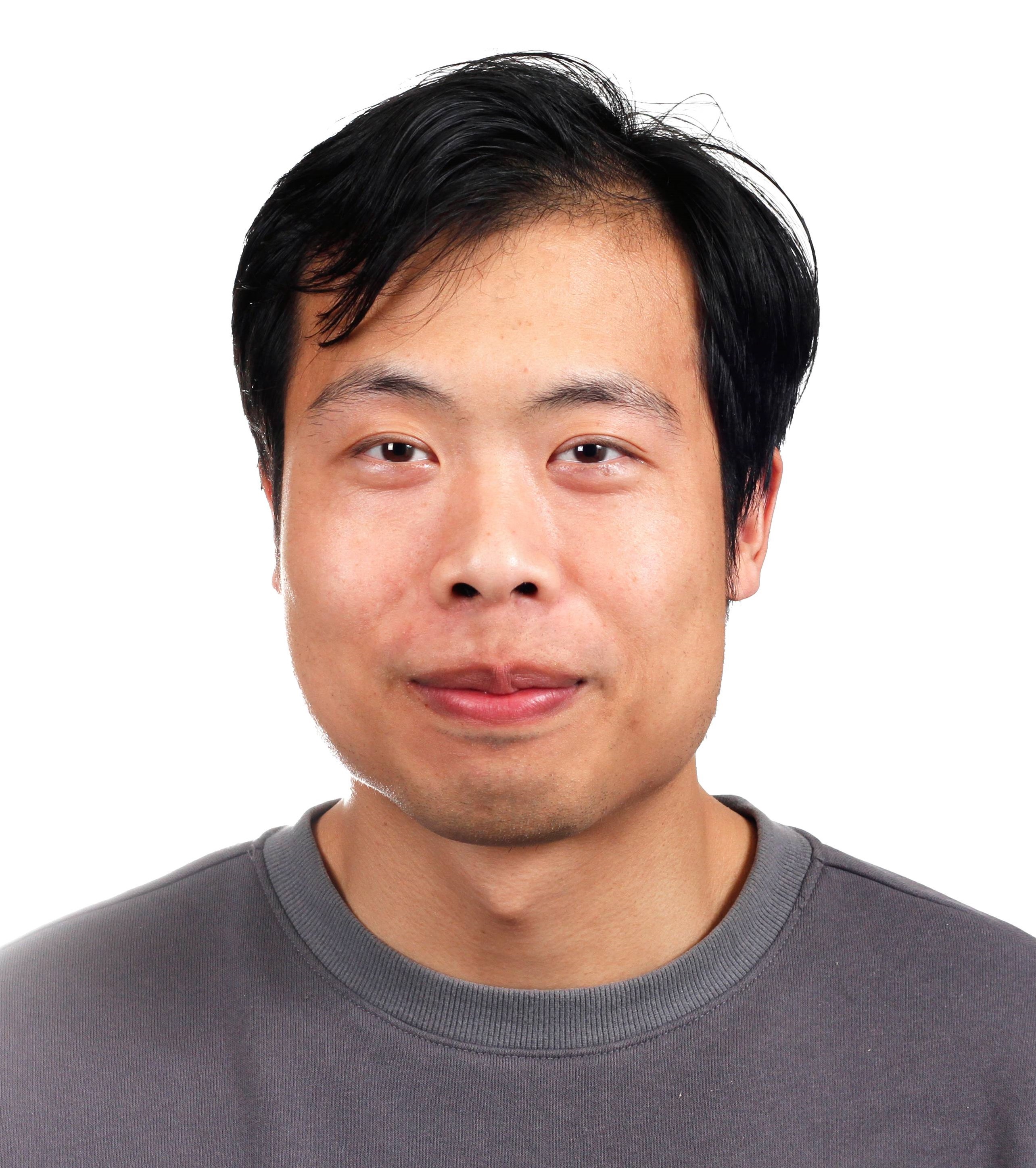}}]{Chuan Huang} (Student member, IEEE) received the B.Sc. from Beihang University in 2018 and the M.Sc. degree from China Electronics Technology Group Corporation Academy of Electronics and Information Technology in 2021. From 2021 to 2014, he studied at Linköping University, Sweden, and he is now a PhD student at the KTH Royal Institute of Technology, Stockholm, Sweden.

His main research interest is magnetic field-based positioning and sensor calibration.
\end{IEEEbiography}

\begin{IEEEbiography}
[{\includegraphics[width=1in,height=1.25in,clip,keepaspectratio]{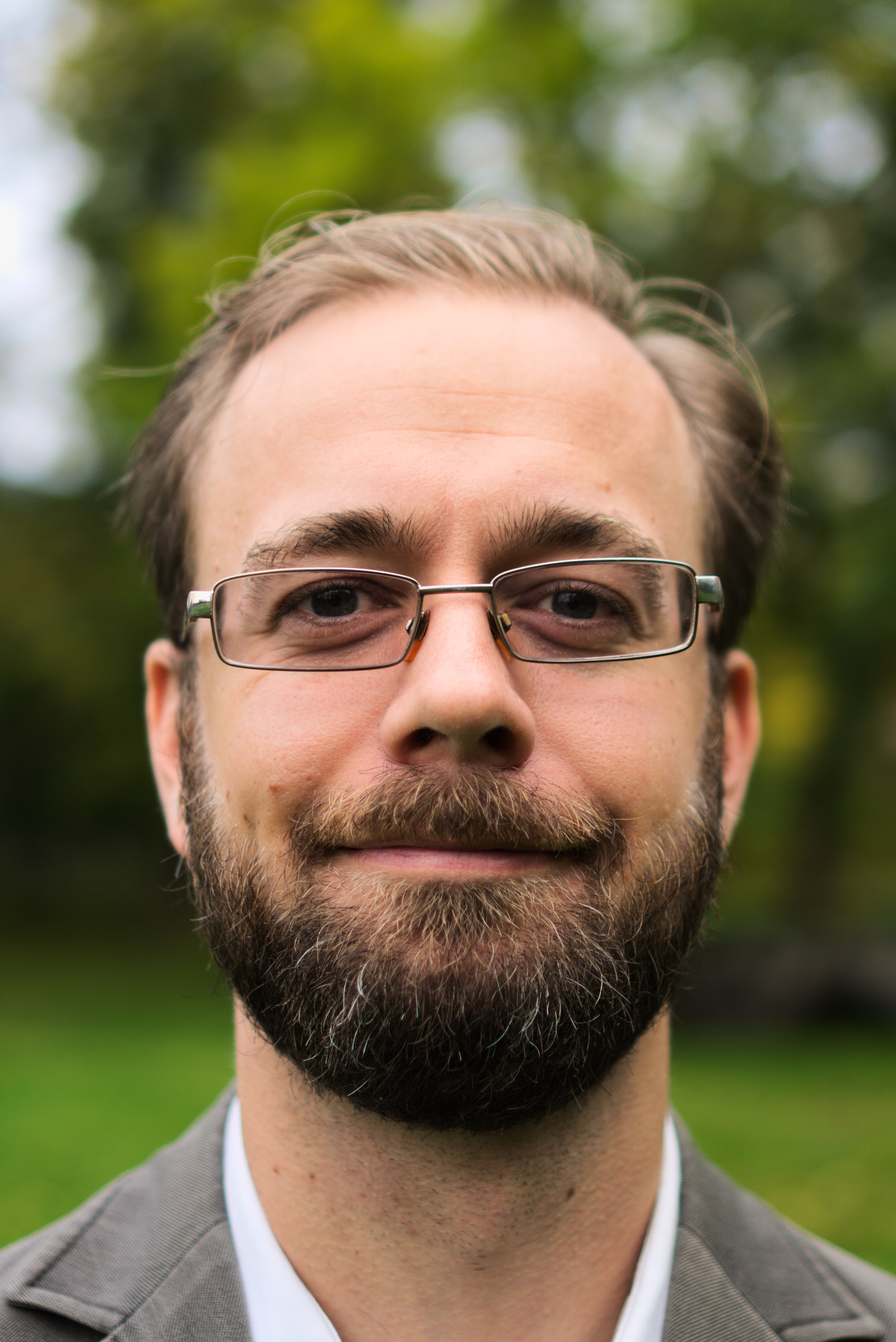}}]{Gustaf Hendeby} (Senior member, IEEE) received the M.Sc. degree in applied physics and electrical engineering in 2002 and the Ph.D. degree in automatic control in 2008, both from Linkoping University, Linkoping, Sweden. He is Associate Professor and Docent in the division of Automatic Control, Department of Electrical Engineering, Linkoping University. 

He worked as Senior Researcher at the German Research Center for Artificial Intelligence (DFKI) 2009–2011, and Senior Scientist at Swedish Defense Research Agency (FOI) and held an adjunct Associate Professor position at Linkoping University 2011–2015. His main research interests are stochastic signal processing and sensor fusion with applications to nonlinear problems, target tracking, and simultaneous localization and mapping (SLAM), and is the author of several published articles and conference papers in the area. He has experience of both theoretical analysis as well as implementation aspects. Dr. Hendeby was an Associate Editor for IEEE Transactions on Aerospace and Electronic Systems in the area of Target Tracking and Multisensor Systems 2018--2025, and is since 2025 a Senior Editor. In 2022 he served as general chair for the 25th IEEE International Conference on Information Fusion (FUSION) in Linkoping, Sweden.
\end{IEEEbiography}

\begin{IEEEbiography}
[{\includegraphics[width=1in,height=1.25in,clip,keepaspectratio]{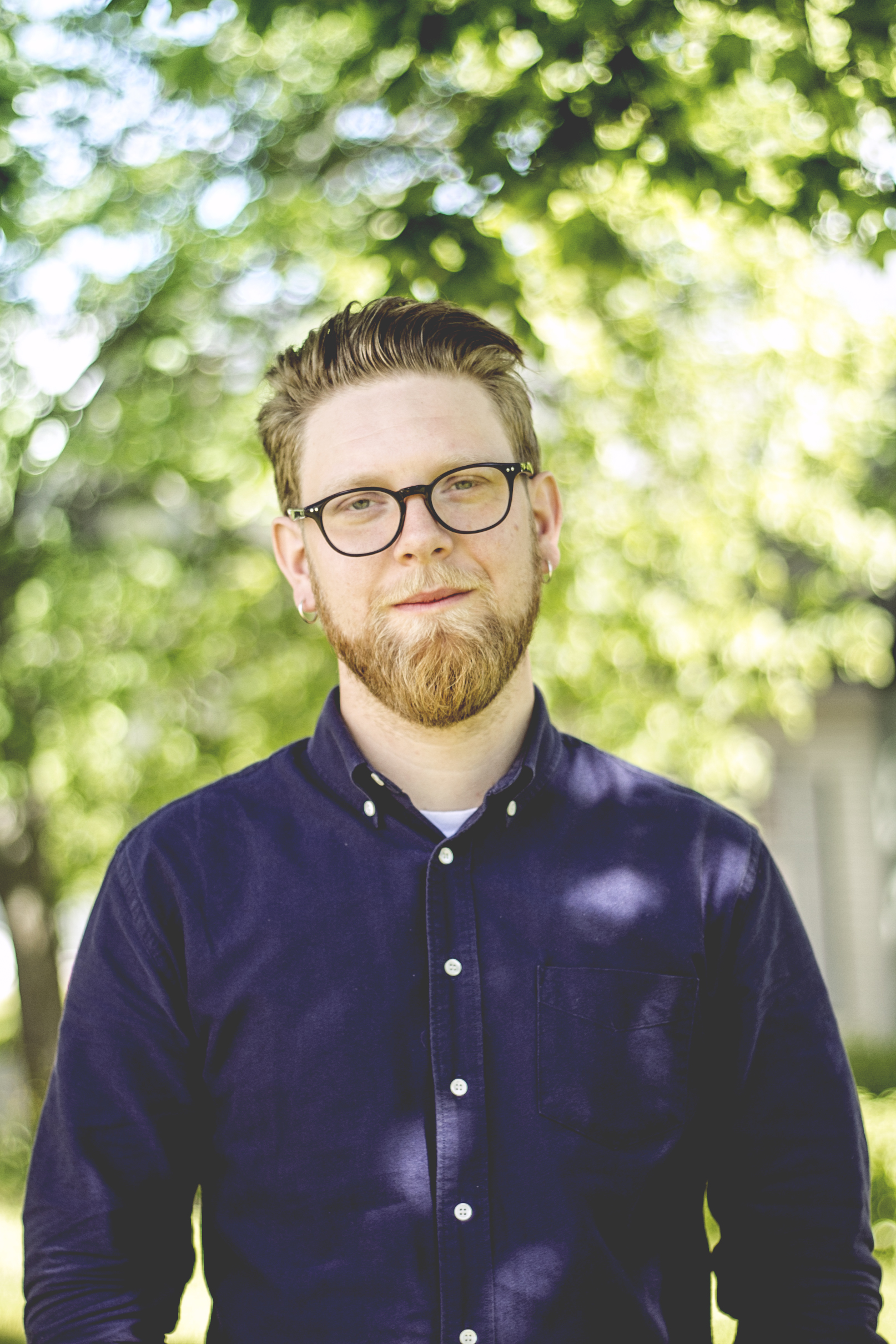}}]{Isaac Skog} (Senior Member, IEEE) received the B.Sc. and M.Sc. degrees in electrical engineering from the KTH Royal Institute of Technology, Stockholm, Sweden, in 2003 and 2005, respectively. In 2010, he received the Ph.D. degree in signal processing with a thesis on low-cost navigation systems. 

In 2009, he spent 5 months with the Mobile Multi-Sensor System Research Team, University of Calgary, Calgary, AB, Canada, as a Visiting Scholar and in 2011 he spent 4 months with the Indian Institute of Science, Bangalore, India, as a Visiting Scholar. Between 2010 and 2017, he was a Researcher with the KTH Royal Institute of Technology. He is currently an Associate Professor with Linköping University, Linköping, Sweden, and a Senior Researcher with Swedish Defence Research Agency (FOI), Stockholm, Sweden. He is the author and coauthor of more than 60 international journal and conference publications. He was the recipient of the Best Survey Paper Award by the IEEE Intelligent Transportation Systems Society in 2013.
\end{IEEEbiography}

\end{document}